\documentclass[10pt,twocolumn,letterpaper]{article}

\usepackage[pagenumbers]{cvpr} 

\usepackage{graphicx}
\usepackage{amsmath}
\usepackage{amssymb}
\usepackage{booktabs}

\usepackage{url}            
\usepackage{booktabs}       
\usepackage{amsfonts}       
\usepackage{nicefrac}       
\usepackage{microtype}      
\usepackage{xcolor,colortbl}
\usepackage{graphicx}
\usepackage{tabularx}
\usepackage{tabulary}
\usepackage{multirow}
\usepackage{subcaption} 
\usepackage{xfrac}
\usepackage{wrapfig}

\usepackage[pagebackref,breaklinks,colorlinks]{hyperref}

\usepackage{wrapfig}

\usepackage{xspace}

\usepackage{listings}
\usepackage{upquote}  
\usepackage{xcolor}
\definecolor{codegreen}{rgb}{0,0.6,0}
\definecolor{codegray}{rgb}{0.5,0.5,0.5}
\definecolor{codepurple}{rgb}{0.58,0,0.82}
\definecolor{backcolour}{rgb}{0.95,0.95,0.92}

\lstdefinestyle{mystyle}{
    backgroundcolor=\color{backcolour},   
    commentstyle=\color{codegreen},
    keywordstyle=\color{magenta},
    numberstyle=\tiny\color{codegray},
    stringstyle=\color{codepurple},
    basicstyle=\ttfamily\footnotesize,
    breakatwhitespace=false,         
    breaklines=true,                 
    captionpos=b,                    
    keepspaces=true,                 
    numbers=left,                    
    numbersep=5pt,                  
    showspaces=false,                
    showstringspaces=false,
    showtabs=false,                  
    tabsize=2
}

\newcommand{\PAR}[1]{\vskip4pt \noindent {\bf #1~}}

\definecolor{todocolor}{rgb}{0.82, 0.41, 0.12}

\definecolor{fixedcolor}{HTML}{333333}
\definecolor{indepcolor}{HTML}{A8165B}
\definecolor{consicolor}{HTML}{66AED2}
\definecolor{fnmatcolor}{HTML}{7ECD31}

\def \shampoo {\includegraphics[height=0.7\baselineskip]{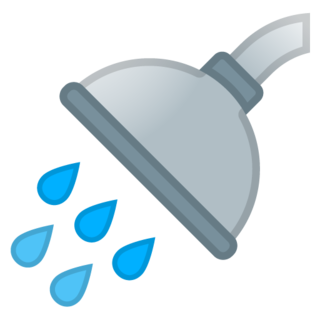}\includegraphics[height=0.7\baselineskip]{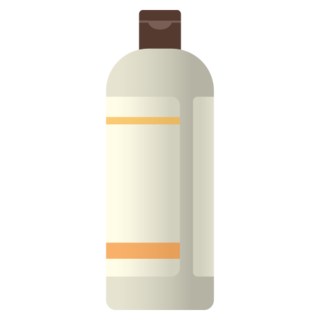}\xspace}

\usepackage[capitalize]{cleveref}
\crefname{section}{Sec.}{Secs.}
\Crefname{section}{Section}{Sections}
\Crefname{table}{Table}{Tables}
\crefname{table}{Tab.}{Tabs.}

\def\cvprPaperID{10832} 
\def\confName{CVPR}
\def\confYear{2022}

\newcommand{\authsep}{\;\;}

\begin{document}

\title{Knowledge distillation:\\
A good teacher is patient and consistent}

\author{Lucas Beyer$^{\star}$ \authsep Xiaohua Zhai$^{\star}$ \authsep Amélie Royer$^{\star\dagger}$ \authsep Larisa Markeeva$^{\star\ddagger}$ \authsep Rohan Anil \authsep Alexander Kolesnikov$^{\star}$ \\
Google Research, Brain Team \\
\texttt{\{lbeyer,xzhai,akolesnikov\}@google.com}}
\maketitle
{\let\thefootnote\relax\footnote{
{$^{\star}$equal contribution \\ $^{\quad\;\;\ \dagger}$ work done at Google, while being a PhD student in IST Austria. \\ $^{\quad\;\;\ \ddagger}$ work done at Google, while being a PhD student in Skoltech.}}}

\begin{abstract}
There is a growing discrepancy in computer vision between large-scale models that achieve state-of-the-art performance and models that are affordable in practical applications. In this paper we address this issue and significantly bridge the gap between these two types of models. Throughout our empirical investigation we do not aim to necessarily propose a new method, but strive to identify a robust and effective recipe for making state-of-the-art large scale models affordable in practice. We demonstrate that, when performed correctly, knowledge distillation can be a powerful tool for reducing the size of large models without compromising their performance. In particular, we uncover that there are certain implicit design choices, which may drastically affect the effectiveness of distillation. Our key contribution is the explicit identification of these design choices, which were not previously articulated in the literature. We back up our findings by a comprehensive empirical study, demonstrate compelling results on a wide range of vision datasets and, in particular, obtain a state-of-the-art ResNet-50 model for ImageNet, which achieves 82.8\% top-1 accuracy.\footnote{We provide code and models in the {\tt big\_vision} codebase~\cite{big_vision}.}  
\end{abstract}

\section{Introduction}\label{sec:intro} 
Large-scale vision models currently dominate many areas of computer vision. Recent state-of-the-art models for image classification~\cite{brain2020bit,EfficientNet,xie2020self,brain2021vit,MLP}, object detection~\cite{ghiasi2020simple,Swin-L} or semantic segmentation~\cite{zoph2020rethinking} push model size to the limits allowed by modern hardware. Despite their impressive performance, these models are rarely used in practice due to high computational costs.
Instead, practitioners typically use much smaller models, such as ResNet-50~\cite{brain2020bit} or MobileNet~\cite{howard2017mobilenets}, which are order(s) of magnitude cheaper to run. 
According to the download counts of five BiT models from \emph{Tensorflow Hub}, the smallest ResNet-50~\cite{he2016deep} model has been downloaded for significantly more times than the larger ones. 
As a result, many recent improvements in vision do not translate to real-world applications. 

\begin{figure}[t]
\centering
  \includegraphics[width=0.85\linewidth]{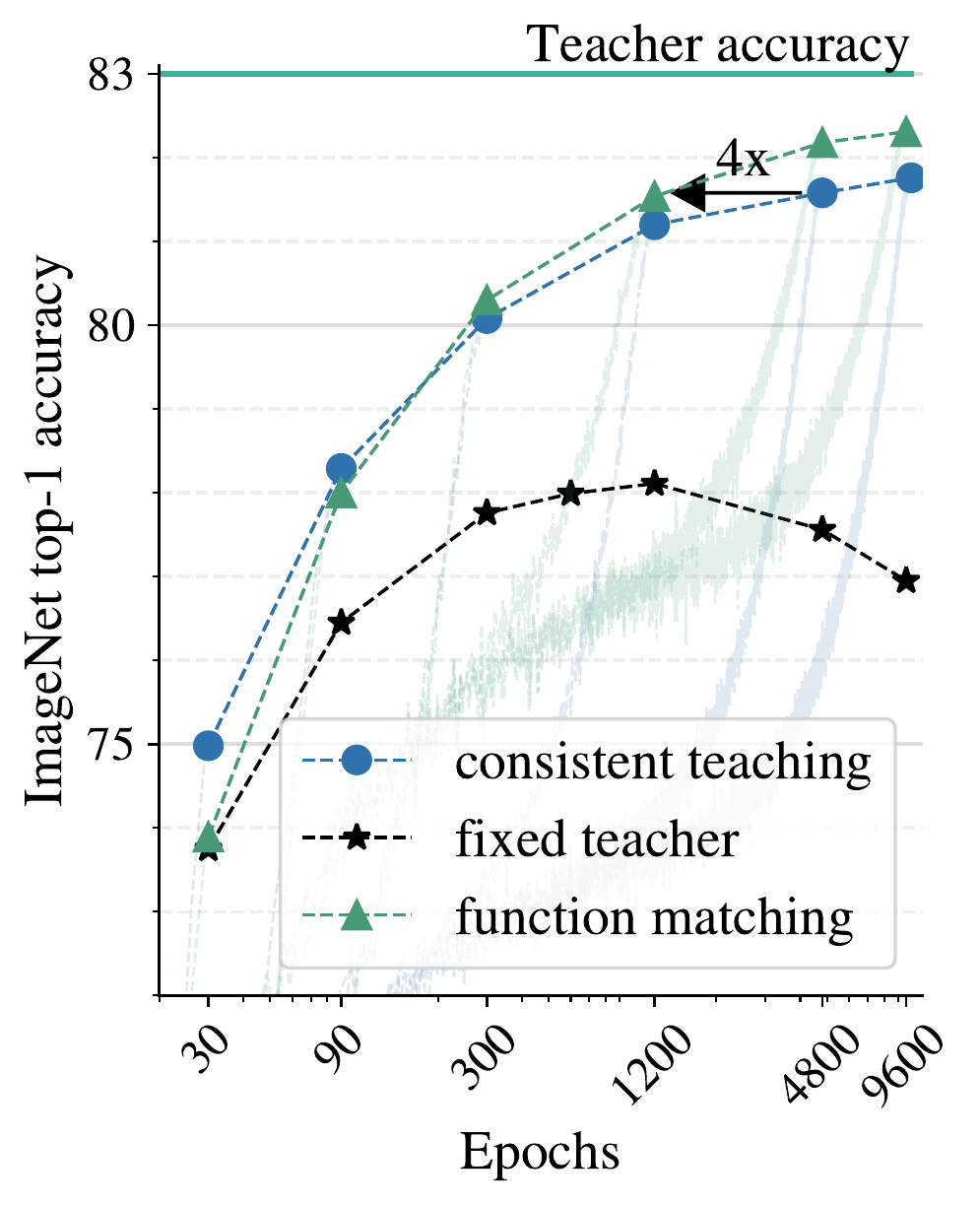}
 \caption{We demonstrate that distillation works the best when we train \textit{patiently} for a large number of epochs and provide \textit{consistent} image views to teacher and student models (green and blue lines). This can be contrasted to a popular setting of distilling with precomputed teacher targets (black line), which works much worse.}
 \end{figure}

To address this problem, we concentrate on the following task: given a specific application and a large model that performs very well on it, we aim to compress the model to a smaller and more efficient architecture without compromising performance. 
There are two widely used paradigms that target this task: model pruning~\cite{janowsky1989pruning} and knowledge distillation~\cite{hinton}. \emph{Model pruning} reduces the large model's size by stripping away its parts. This procedure can be restrictive in practice:
first, it does not allow changing the model family, say from a ResNet to a MobileNet.
Second, there may be architecture-dependent challenges, \eg if the model uses group normalization~\cite{wu2018group}, pruning channels may result in the need to dynamically re-balance channel groups.

\begin{figure*}[t]%
    \centering
    \includegraphics[width=0.99\textwidth]{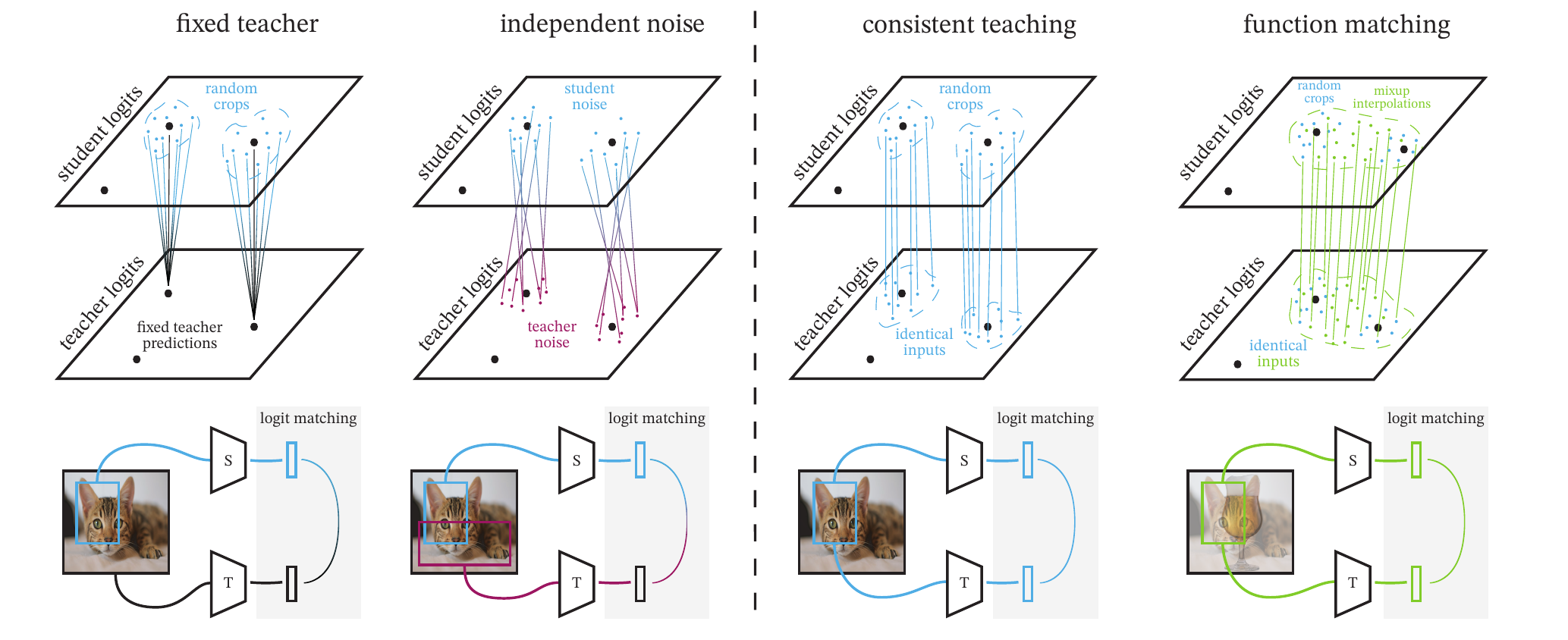}%
    \vspace{-1em}
    \caption{Schematic illustrations of various design choices when doing knowledge distillation. \textcolor{fixedcolor}{\textbf{Left}}: \emph{Teacher} receives a fixed image, while \emph{student} receives a random augmentation. \textcolor{indepcolor}{\textbf{Center-left}}: \emph{Teacher} and \emph{student} receive independent image augmentations. \textcolor{consicolor}{\textbf{Center-right}}: \emph{Teacher} and \emph{student} receive consistent image augmentations. \textcolor{fnmatcolor}{\textbf{Right}}: \emph{Teacher} and \emph{student} receive consistent image augmentations plus the input image manifold is extended by including linear segments between pairs of images (known as \emph{mixup}~\cite{zhang2018mixup} augmentation).}%
    \vspace{-1em}
    \label{fig:method}%
\end{figure*}

Instead, we concentrate on the \emph{knowledge distillation} approach which does not suffer from these drawbacks.
The idea behind knowledge distillation is to ``distill'' a \textit{teacher model}, in our case a large and cumbersome model or ensemble of models, into a small and efficient \textit{student model}.
This works by forcing the student's predictions (or internal activations) to match those of the \emph{teacher}, thus naturally allowing a change in the model family as part of compression.
We closely follow the original distillation setup from~\cite{hinton} and find it surprisingly effective when done right: We interpret distillation as a task of matching the \emph{functions} implemented by the teacher and student, as illustrated in Figure~\ref{fig:method}. With this interpretation, we discover two principles of knowledge distillation for model compression. First, teacher and student should process the exact same input image views or, more specifically, same crop and augmentations. Second, we want the functions to match on a large number of support points to generalize well. Using an aggressive variant of mixup~\cite{zhang2018mixup}, we can generate support points outside the original image manifold. With this in mind, we experimentally demonstrate that consistent image views, aggressive augmentations and very long training schedules are the key to make \emph{model compression via knowledge distillation} work well in practice.

Despite the apparent simplicity of our findings, there are multiple reasons that may commonly prevent researchers (and practitioners) from making the design choices that we suggest.
\textbf{First}, it is tempting to precompute the teacher's activations for an image offline once to save compute, especially for very large teachers. As we will show, this fixed teacher approach does not work well.
\textbf{Second}, knowledge distillation is also commonly used in different contexts (other than model compression), where authors recommend different or even opposite design choices~\cite{xie2020self,tarvainen2017mean,yun2021re}, see Figure~\ref{fig:method}.
\textbf{Third}, knowledge distillation requires an atypically large number of epochs to reach best performance, much more than commonly used for supervised training.   
Finally, choices which may look suboptimal in training of regular length often end up being best for long runs, and vice-versa. 
 
In our empirical study, we mostly concentrate on compressing the large BiT-ResNet-152x2 from~\cite{brain2020bit} that was pretrained on the ImageNet-21k dataset~\cite{russakovsky2015imagenet} and fine-tuned to the relevant datasets of interest. We distill it to a standard ResNet-50 architecture~\cite{he2016deep} (but replace batch normalization with group normalization) on a range of small and mid-sized datasets without compromising accuracy.
We also achieve very strong results on the ImageNet~\cite{imagenet} dataset: with a total number of 9600 epochs for distillation, we set the new ResNet-50 SOTA 82.8\% on ImageNet. This is 4.4\% better than the ResNet-50 model from~\cite{brain2020bit}, and 2.2\% better than the best ResNet-50 model in the literature, which uses a more complex setup~\cite{shen2020mealv2}.
Finally, we demonstrate that our distillation recipe also works when simultaneously compressing and changing the model family, \eg BiT-ResNet architecture to the MobileNet architecture.

\section{Experimental setup}\label{sec:setup}

In this section, we introduce the experimental setup and benchmarks we use throughout the paper. Given a large-scale vision model (the \emph{teacher}, or T) with high accuracy on a particular task, we aim to compress this model to a much smaller model (the \emph{student}, or S) without compromising its performance. 
Our compression recipe relies on knowledge distillation, as introduced in~\cite{hinton},  and a careful investigation of several key ingredients in the training setup.

\label{datasets_introduction}
\textbf{Datasets, metrics and evaluation protocol.} We conduct experiments on five popular image classification datasets: \emph{flowers102}~\cite{dataflowers}, \emph{pets}~\cite{datapets},  \emph{food101}~\cite{datafood}, \emph{sun397}~\cite{datasun} and \emph{ILSVRC-2012} (``ImageNet'')~\cite{imagenet}. 
These datasets span diverse image classification scenarios; In particular, they vary in the number of classes, from $37$ to $1000$ classes, and total number of training images, from $1020$ to $1281167$ training images.
This allows us to verify our distillation recipe for a broad range of practical settings and ensure its robustness.

As a metric, we always report classification accuracy. For all datasets, we perform design choices and hyperparameters selection using a \emph{validation} split, and report final results on the \emph{test} set. These splits are defined in the appendix~\ref{app:splits}.   

\textbf{Teacher and student models.} Throughout the paper, we opt for using  pre-trained teacher models from BiT~\cite{brain2020bit}, which provides a large collection of ResNet models pretrained on ILSVRC-2012 and ImageNet-21k datasets, with state-of-the-art accuracy. The only significant differences  between BiT-ResNets and standard ResNets is their use of group normalization layer~\cite{wu2018group} and weight standardization~\cite{qiao2019micro}, which are used instead of batch normalization~\cite{ioffe2015batch}.

In particular, we concentrate on the BiT-M-R152x2 architecture: a BiT-ResNet-152x2 
(152 layers, `x2' indicates the width multiplier) pretrained on ImageNet-21k.  This model demonstrates excellent performance on a variety of vision benchmarks and it is still manageable to run extensive ablation studies with it. 
It is highly expensive to deploy (requires roughly 10x more compute than the standard ResNet-50), and thus effective compression of this model is of practical importance. 
For the student's architecture, we use a BiT-ResNet-50 variant, referred to as ResNet-50 for brevity.

\textbf{Distillation loss.} We use the KL-divergence between the teacher's $p_t$, and the student's $p_s$ predicted class probability vectors as a distillation loss, as was originally introduced in~\cite{hinton}. We do not use any additional loss term with respect to the original dataset's hard labels: 
\begin{equation}
\mathrm{KL}(p_t || p_s) = \sum\limits_{i \in \mathcal{C}} \left[-p_{t,i} \log p_{s,i} + p_{t,i} \log p_{t,i} \right],
\end{equation}
where $\mathcal{C}$ is a set of classes. Also, as in \cite{hinton}, we introduce a temperature parameter $T$, which is used to adjust the entropy of the predicted softmax-probability distributions before they are used in the loss computation: $p_s \propto \exp(\frac{\log p_s}{T})$ and $p_t \propto \exp(\frac{\log p_t}{T})$.

\begin{figure*}[t]%
    \centering
    \includegraphics[width=\textwidth]{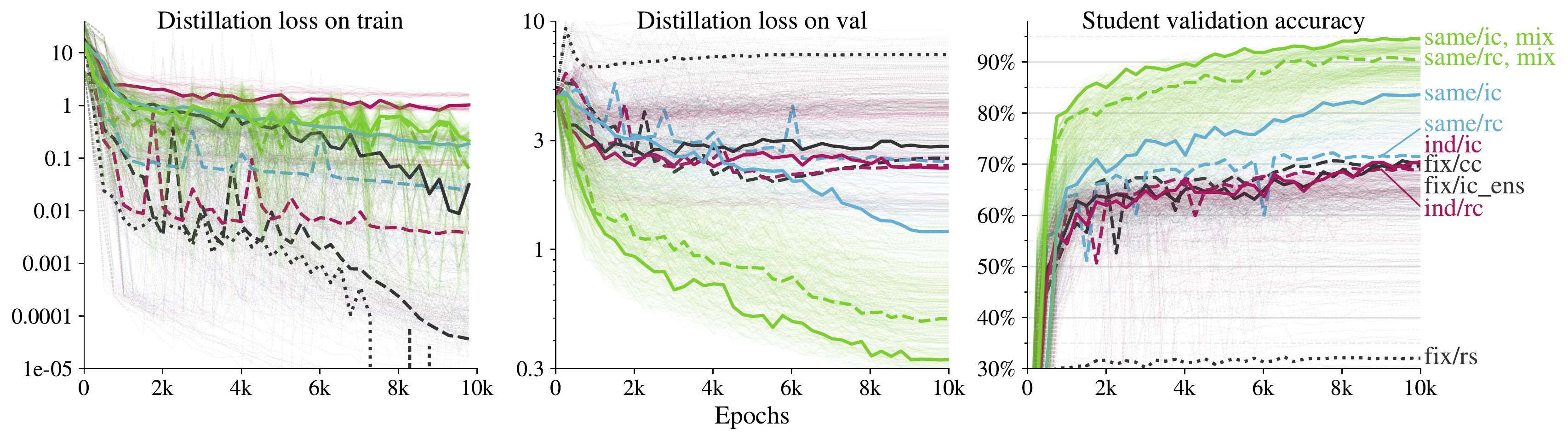}%
    \caption{Experimental validation of the ``consistency'' requirement on the \emph{Flowers102} dataset. Colors match different knowledge distillation design choices as introduced in Figure~\ref{fig:method} and Section \ref{sec:designchoices}. Note that while the \textcolor{fixedcolor}{fixed teacher} settings achieve significantly lower distillation loss, they lead to students which do not generalize  well. In contrast, \textcolor{consicolor}{consistent teaching} and \textcolor{fnmatcolor}{function matching} approaches lead to significantly higher student performance. Similar results on more datasets are reported in Appendix~\ref{sec:app:consistency}.}%
    \label{fig:consistency}%
\end{figure*}

\textbf{Training setup.} For optimization, we train our models with the Adam optimizer~\cite{adam} with default parameters, except for the initial learning rate that is part of our hyperparameter exploration. 
We use a cosine learning rate schedule~\cite{loshchilov2016sgdr} without warm restarts. 
We also sweep over the weight decay loss coefficient for all our experiments (for which we use a ``decoupled'' weight decay convention~\cite{loshchilov2017decoupled}).
To stabilize training we enable gradient clipping with a threshold of 1.0 on the global L2-norm of a gradient. Finally, we use batch size 512 for all our experiments, except for models trained on ImageNet, where we train with  batch size 4096.
For the remaining hyperparameters, we discuss their sweeping range together with corresponding experiments in the next section.

One additional important component of our recipe is the  \emph{mixup} data augmentation strategy~\cite{zhang2018mixup}.
In particular, we introduce a mixup variant in our ``function matching'' strategy (see Section \ref{par:funcmatching}), in which we 
use ``agressive'' mixing coefficients sampled uniformly from $[0, 1]$, which can be seen as an extreme case of 
the originally proposed sampling from $\beta$-distribution.
  
Unless explicitly specified otherwise, for prepossessing we use an ``inception-style'' crop~\cite{szegedy2015going} and then resize images to a fixed square size. 
Furthermore, in order to make our extensive analysis computationally feasible (overall we trained dozens of thousands of models), we use relatively low input resolution and resize input images to $128 \times 128$ size, except for our ImageNet experiments, that use the standard input $224 \times 224$ resolution.

For all our experiments we use Google Cloud TPU accelerators~\cite{jouppi2017datacenter}. We also report our batch sizes, epochs or total number of update steps, which allow to estimate resource requirements for any particular experiment of interest. Code and checkpoints are made publicly available~\cite{big_vision}.

\section{Distillation for model compression}\label{sec:exp}

\subsection{Investigating the "consistent and patient teacher" hypothesis}
\label{sec:hpsweep}

In this section, we perform an experimental verification of our hypothesis formulated in the introduction and visualised in Figure~\ref{fig:method}, 
 that distillation works best when seen as function matching, \ie when the  student and teacher see consistent views of the input images, synthetically "filled" via mixup, and when \emph{student is trained} using long training schedule (i.e. the ``teacher'' is patient).

To make sure that our findings are robust, we perform a very thorough analysis on four small and medium scale datasets, namely \emph{Flowers102}~\cite{dataflowers} (1020 training images), \emph{Pets}~\cite{datapets} (3312 training images), \emph{Food101}~\cite{datafood} (about 68k training images), and \emph{SUN397}~\cite{datasun} (76k training images).

In an effort to remove any confounding factors, for each individual distillation setting we sweep over all combinations of learning rates $\{0.0003, 0.001, 0.003, 0.01\}$, weight decays $\{1\cdot10^{-5}, 3\cdot10^{-5}, 1\cdot10^{-4}, 3\cdot10^{-4}, 1\cdot10^{-3}\}$, and distillation temperatures $\{1, 2, 5, 10\}$.
In all reported figures, we show every single run as a low opacity curve, and highlight the one with the best final validation accuracy.
We provide corresponding test accuracies in Appendix~\ref{sec:app:table}.

\subsubsection{Importance of ``consistent'' teaching}
\label{sec:designchoices}

\begin{figure*}[t]%
    \centering
    \includegraphics[width=\textwidth]{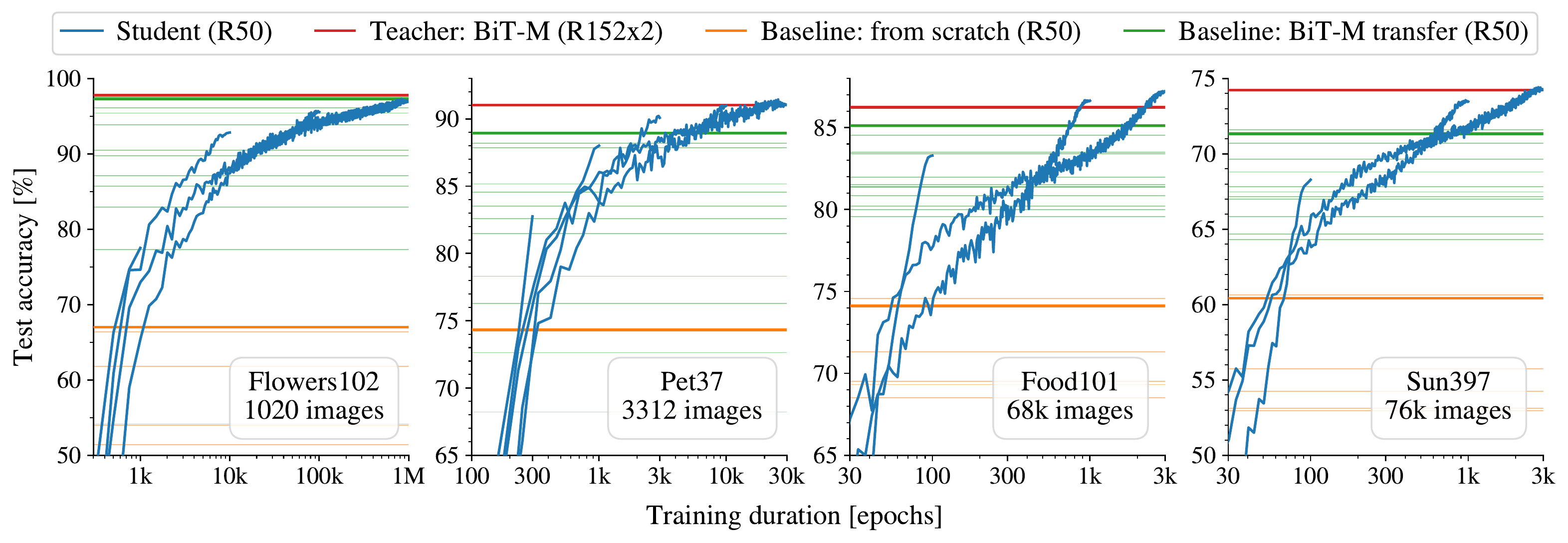}%
    \caption{One needs patience along with consistency when doing distillation. Eventually, the teacher will be matched; this is true across various datasets of different scale.}
    \label{fig:patience}%
\end{figure*}

First, we demonstrate that the consistency criterion, \ie student and teacher seeing the same views, is the only way of performing distillation which reaches peak student performance across all datasets consistently.
For this study, we define multiple distillation configurations which correspond to instantiations of all four options sketched in Figure~\ref{fig:method}, with the same color coding:
\begin{itemize}
    \item \textcolor{fixedcolor}{\bf Fixed teacher.} We explore several options where the teacher's predictions are constant for a given image (precomputed target).
    The simplest (and worst) method is {\tt fix/rs}, where the image is just resized to $224^2$px for both student and teacher. {\tt fix/cc} follows a more common approach of using a fixed central crop for the teacher and a mild random crop for the student.
    {\tt fix/ic\_ens} is a heavy data augmentation approach where the teacher's prediction is the average of 1k inception crops, which we verified to improve the teacher's performance. The student also uses random inception crops. The two latter settings are similar to the input noise strategy from the ``noisy student'' paper~\cite{xie2020self}.
    \item \textcolor{indepcolor}{\bf Independent noise.} We instantiate this common strategy in two ways: {\tt ind/rc} computes two independent mild random crops for the teacher and student respectively, while {\tt ind/ic} uses the heavier inception crop instead. A similar setup was used in~\cite{tarvainen2017mean}.
    \item \textcolor{consicolor}{\bf Consistent teaching.} In this approach, we randomly crop the image only once, either with mild random cropping ({\tt same/rc}) or heavy inception crop ({\tt same/ic}), and use this same crop for the input to both the student and the teacher.
    \item \textcolor{fnmatcolor}{\bf Function matching.} This approach extends consistent teaching, by expanding an input manifold of images through mixup ({\tt mix}), and, again, providing consistent inputs to the student and the teacher. \label{par:funcmatching}
    For brevity, we sometimes refer to this approach as ``FunMatch''.
\end{itemize}

Figure~\ref{fig:consistency} shows 10\,000 epoch training curves on \emph{Flowers102} dataset in all of these configurations.
These results clearly show that ``consistency'' is the key: all ``inconsistent'' distillation settings plateau at a lower score, while consistent settings increase student performance significantly, with the function matching approach working the best.
Furthermore, the training losses show that, for such small datasets, using a fixed teacher leads to strong overfitting. In contrast, function matching never reaches such loss on the training set while generalizing much better to the validation set.
Due to space constraints, we show analogous results for other datasets and training durations in Appendix~\ref{sec:app:consistency}.

\subsubsection{Importance of ``patient'' teaching}

One can interpret distillation as a variant of supervised learning, where labels (potentially soft) are provided by a strong \emph{teacher} model. This is especially true when the \emph{teacher} predictions are (pre)computed for a single image view. This approach inherits all problems of the standard supervised learning, e.g. aggressive data augmentations may distort actual image label, while less aggressive augmentations may cause overfitting.

However, things change if we interpret distillation as function matching, and, crucially, make sure to provide consistent inputs to the \emph{student} and \emph{teacher}. In this case we can be very aggressive with image augmentations: even if an image view is too distorted, we still will make a progress towards matching the relevant functions on this input. Thus, we can be more opportunistic with augmentations and avoid overfitting by doing aggressive image augmentations and, if true, optimize for very long time until the \emph{student}'s function comes close to the \emph{teacher}'s. 

We empirically confirm our intuition in Figure~\ref{fig:patience}, where for each dataset we show the evolution of test accuracy during training of the best function matching student (according to validation), for different amounts of training epochs.
The teacher is shown as a red line and is always reached eventually, after a much larger number of epochs than one would ever use in a supervised training setup.
Crucially, there is no overfitting even when we optimize for a 1M epochs.

We also trained and tuned two more baselines for reference: training a ResNet-50 from scratch using the dataset original hard labels, as well as transferring a ResNet-50 that was pre-trained on ImageNet-21k.
For both of these baselines, we heavily tune learning rate and weight decay as described in Section~\ref{sec:hpsweep}.
The model trained from scratch using the original labels is substantially outperformed by our student.
The transfer model fares much better, but is eventually also outperformed.

Notably, training for a relatively short but common duration of 100 epochs leads to much worse performance than the transfer baseline.
Overall, the ResNet-50 student patiently and consistently matches the very strong but much more expensive ResNet-152x2 teacher across the board.

\begin{figure*}[t]
\centering
  \includegraphics[width=0.325\linewidth]{figs/schedule.pdf}%
  \quad
  \includegraphics[width=0.3\linewidth]{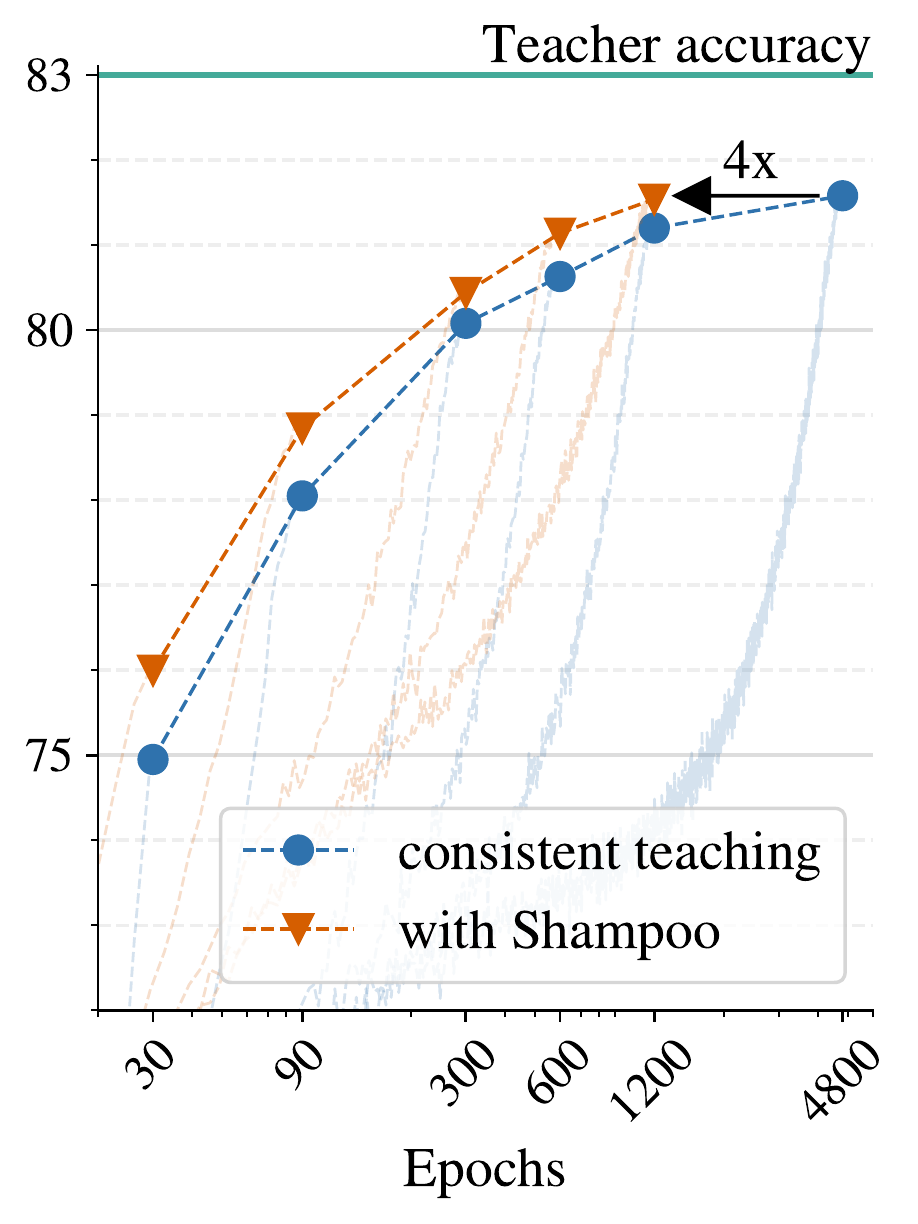}%
  \quad
  \includegraphics[width=0.3\linewidth]{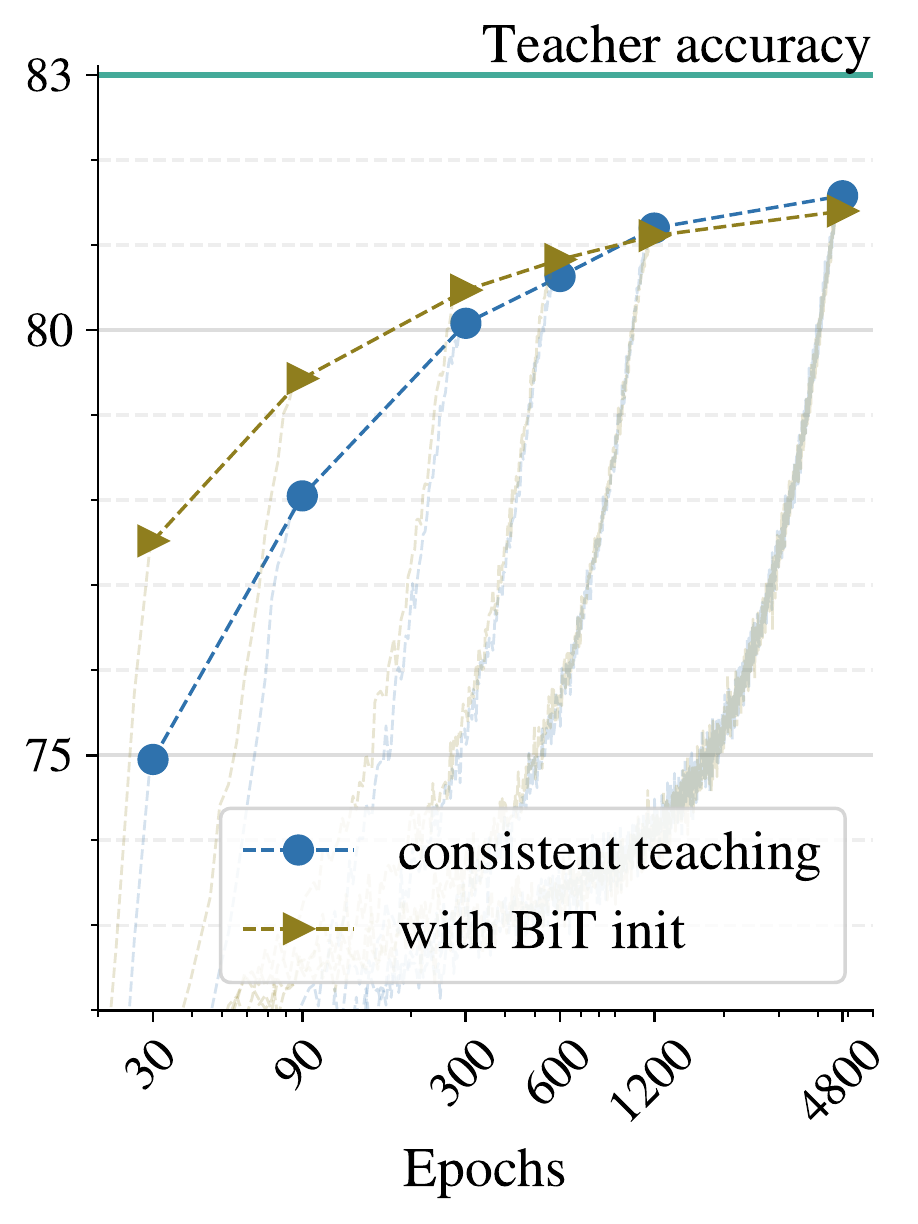}
  \captionof{figure}{
  \textbf{Left}: Top-1 accuracy on ImageNet of three distillation setups: (1) \textcolor{fixedcolor}{fixed teacher}; (2) \textcolor{consicolor}{consistent teaching}; (3) \textcolor{fnmatcolor}{function matching} (``FunMatch'').
  Light color curves show accuracy throughout training, while the solid scatter plots are the final results.
  The student with a fixed teacher eventually saturates and overfits to it.
  Both consistent teaching and function matching do not exhibit overfitting or saturation.
  \textbf{Middle}: Reducing the optimization cost, via \emph{Shampoo}~\protect\shampoo preconditioning; with 1200 epochs, it is able to match the baseline trained for 4800 epochs.
  \textbf{Right}: Initializing \emph{student} with pre-trained weights improves short training runs, but harms for the longest schedules.}
  \label{fig:imagenet}
\end{figure*}

\subsection{Scaling up to ImageNet}

Based on our insights from the previous sections, 
we now investigate how the proposed distillation recipe scales to the widely used and more challenging ImageNet dataset \cite{imagenet}. Following the same protocol as before, in Figure~\ref{fig:imagenet}~(left), we report student accuracy curves throughout training for three distillation settings: 
(1) fixed teacher, (2) consistent teaching and (3) function matching.
For reference, our base teacher model reaches a top-1 accuracy of 83.0\%.
Fixed teacher again suffers from long training schedules, and starts overfitting after 600 epochs.
In contrast, the consistent teaching approaches continuously improves performance as the training duration increases.
From this we can conclude that consistency is a key to make distillation work on ImageNet, similar to the behaviors on the previously discussed small and mid-sized datasets.

Compared to simple consistent teaching, function matching performs slightly worse with short schedules, which likely happens due to underfitting. But when we increase the length of training schedule, the improvement of function matching becomes apparent: for instance with only 1200 epochs, it is able to match the performance of consistent teaching at 4800 epochs, thus saving 75\% compute resource.
Finally, for the longest run of function matching we experimented on, the vanilla \textbf{ResNet-50}  student architecture achieves \textbf{82.31\%} top-1 accuracy on ImageNet. 

\subsection{Distilling across different input resolutions}

So far, we have assumed that both the \emph{student} and \emph{teacher} receive the same standard input resolution of 224px.
However, it is possible to pass images of different resolution to the student and the teacher, while still being consistent: one simply has to perform the crop on the original high-resolution image, and subsequently resize it differently for the student and the teacher: their views will be consistent, albeit at different resolutions.
This insight can be leveraged for learning from a better, higher resolution, teacher~\cite{brain2020bit,touvron2019FixRes}, but also for training a smaller, faster student~\cite{bello2021revisiting}.
We investigate both directions: first, following~\cite{bello2021revisiting}, we train a ResNet-50 student with an input resolution of 160px while leaving the teacher resolution unchanged (224px). 
This results in a twice faster model, which still achieves remarkable 80.49\% top-1 accuracy (see Table~\ref{tbl:inputres}), compared to the best published  78.8\% at this resolution using an array of modifications~\cite{bello2021revisiting}.

Second, following~\cite{brain2020bit}, we distill a teacher that was fine-tuned at a resolution of 384px (and attains 83.7\% top-1 accuracy), this time leaving the student resolution unchanged, \ie consuming a 224px input image.
Compared to the baseline teacher, this provides a modest but consistent
improvement across the board, as shown in Table~\ref{tbl:inputres}.

\subsection{Optimization: A second order preconditioner (\protect\shampoo) improves training efficiency}
\label{sec:limitations1}

We observe that optimization efficacy creates a computational bottleneck for our distillation recipe with ``function matching'' perspective due to long training schedules. Intuitively, we believe that optimization difficulties stem from the fact that it is much harder to fit a general \emph{function} with multivariate outputs, rather than fixed image-level labels. Thus, we conduct an initial exploration, whether more powerful optimizers can do a much better job at our task.

To this end, we change the underlying optimizer from Adam to Shampoo~\cite{anil2021scalable}, with the second order preconditioner. In Figure~\ref{fig:imagenet}~(middle) we observe that Shampoo achieves the same test accuracy reached by Adam at 4800 epochs in just 1200 epochs, and with minimal step time overhead. And, in general, we observe consistent improvement over Adam in all our experimental settings. Experimental details on the Shampoo optimizer are provided in the Appendix~\ref{sec:app:shampoo}. 

\subsection{Optimization: A good initialization improves short runs but eventually falls behind}
\label{sec:limitations2}

Motivated by transfer learning literature~\cite{he2018rethinking, brain2020bit} and~\cite{shen2020mealv2}, where a good initialization is able to significantly shorten the training cost and achieve a better solution, we try to initialize the student model with a pre-trained BiT-M-ResNet50 weights and show the results in Figure~\ref{fig:imagenet}~(right).

The BiT-M initialization improves more than 2\% when the distillation duration is short.
However, the gap closes when the training schedule is long enough. 
Our observation is similar to the conclusion of~\cite{he2018rethinking}. Starting from 1200 epochs, distilling from scratch matches the BiT-M initializated student, and slightly overtakes it for 4800 epochs. 

\subsection{Distilling across different model families}

Going beyond using different input resolutions for student and teacher, nothing in principle prevents us from using architectures of different families entirely, as our consistent patient teacher approach still applies in this setting. This allows us to efficiently transfer knowledge from stronger and more complex teachers, e.g. ensembles, while keeping the simple architecture of a ResNet50 student, but also transfer the state-of-the-art performance of large ResNet models to more efficient architectures e.g. MobileNet.
We demonstrate this via two experiments.
First, we use an ensemble of two models as teacher and show that this further improves performance.
Second, we train a MobileNet~v3~\cite{howard2019searching} student and obtain the best reported MobileNet~v3 model to date.

\PAR{MobileNet student.} We use MobileNet v3 (Large) as a student, for most experiments we opt for the variant which uses GroupNorm (with the default of 8 groups) instead of BatchNorm.
We do not use any of the training tricks used in the original paper,%
we simply perform function matching.
Our student reaches 74.60\% after 300 epochs, and 76.31\% after 1200 epochs, resulting in the best published MobileNet v3 model. More results are in the Appendix~\ref{sec:app:table}.

\PAR{Ensemble teacher.} We now try a better teacher: we create a model which consists of averaging the logits from our default teacher at 224px resolution, and our teacher at 384px resolution from the previous section. 
This is a different, though closely related, type of teacher which is significantly more powerful but also slower.
This teacher's student is better than our default teacher's student at every duration we tried (Appendix~\ref{sec:app:table}) and, after 9600 epochs, reaches a new state-of-the-art \textbf{top-1 ImageNet accuracy of 82.82\%}.

\begin{table}[t]
  \setlength{\tabcolsep}{0pt}
  \setlength{\extrarowheight}{5pt}
  \renewcommand{\arraystretch}{0.75}
  \newcolumntype{C}{>{\centering\arraybackslash}X}
  \newcolumntype{R}{>{\raggedleft\arraybackslash}X}
  \centering
  \captionof{table}{Top-1 test accuracy for different teacher/student input resolutions (rows) and number of training epochs (columns).}\label{tbl:inputres}
  \vspace{0.5mm}
  \begin{tabularx}{\linewidth}{p{2.3cm}p{0.1cm}Cp{0.01cm}Cp{0.01cm}Cp{0.01cm}C}
    \toprule[1pt]
    \bf{Experiment} && \bf{300} && \bf{1200} && \bf{4800} && \bf{9600} \\
    \midrule
T224 $\rightarrow$ S224 && 80.30 && 81.54 && 82.18 && 82.31 \\
    \midrule
T224 $\rightarrow$ S160 && 78.17 && 79.61 && N/A && 80.49 \\
    \arrayrulecolor{lightgray}\midrule[0.25pt]\arrayrulecolor{black}
T384 $\rightarrow$ S224 && 80.46 && 81.82 && 82.33 && 82.64 \\
    \bottomrule
  \end{tabularx}
\end{table}

\subsection{Comparison to the results from literature.} 

Now, when we introduced our key experiments, we compare our best ResNet-50 models to the best ResNet-50 models available in the literature, see Table~\ref{tbl:sota}. In particular, for $224 \times 224$ input resolution we compare against the original ResNet-50 model from~\cite{he2016deep}, BiT-M-ResNet-50 pretrained on ImageNet-21k dataset~\cite{russakovsky2015imagenet} and previous state-of-the-art model from~\cite{shen2020mealv2}. For $160 \times 160$ input resolution we compare against very recent and competitive model from~\cite{bello2021revisiting}. We observe that our distillation recipe leads the state-of-the-art performance in both cases and by a significant margin.

\subsection{Distilling on the "out-of-domain" data}

By looking at knowledge distillation as ``function matching'', one can draw a reasonable hypothesis that distillation can be done on arbitrary image inputs. In this section we investigate this hypothesis.

We conduct experiments on \emph{pets} and \emph{sun397} datasets. We use our distillation recipe to distill \emph{pets} and \emph{sun397} models using out-of-domain images from the \emph{food101} and \emph{ImageNet} datasets and, for the reference results, also run distillation with ``in-domain'' images from \emph{pets} and \emph{sun397} datasets.

Figure~\ref{fig:wrong_data} summarizes our results. First we observe that distilling using in-domain data works the best. Somewhat surprisingly, even if the images are completely unrelated, distillation still works to some extent, though results get worse. This, for example, means that the \emph{student} model can learn to classify pets with roughly 30\% accuracy by only seeing food images (softly) labeled as breeds of pets. 
Finally, if distillation images are somewhat related or overlapping with the actual ``in-domain'' images (e.g. Pets and ImageNet, or sun397 and ImageNet), then results can be as good (or almost as good) as using ``in-domain'' data, but extra long optimization schedule may be required. 

\begin{table}[t]
  \setlength{\tabcolsep}{0pt}
  \setlength{\extrarowheight}{5pt}
  \renewcommand{\arraystretch}{0.75}
  \newcolumntype{C}{>{\centering\arraybackslash}X}
  \newcolumntype{R}{>{\raggedleft\arraybackslash}X}
  \centering
  \captionof{table}{Comparison of our best and literature ResNet models. The metric is accuracy on ImageNet test split (officially \emph{val} split).}\label{tbl:sota}
  \begin{tabularx}{\linewidth}{p{4.0cm}p{0.8cm}p{0.4cm}cp{0.4cm}c}
    \toprule[1pt]
    \textbf{Model} & Arch. && Res. && Accuracy \\
    \bottomrule
    ``Revisiting ResNet''~\cite{bello2021revisiting} & R50 && 160  && 78.8\%  \\
    \textbf{FunMatch} (T224) & R50 && 160 && \textbf{80.5\%}  \\
    \midrule
    Original ResNet~\cite{he2016deep} & R50  && 224  && 77.2\%  \\
    BiT-M-R50~\cite{brain2020bit} & R50  && 224  && 78.4\%  \\
    Meal-v2~\cite{shen2020mealv2} & R50  && 224  && 80.7\%  \\
    \textbf{FunMatch} (T384+224) & R50 && 224  && \textbf{82.8\%}  \\
    \midrule
    \textcolor{gray}{``Revisiting ResNet''~\cite{bello2021revisiting}} & \textcolor{gray}{R152} && \textcolor{gray}{224} && \textcolor{gray}{82.8\%}  \\
    \bottomrule
  \end{tabularx}
\end{table}

\subsection{Finetuning ResNet-50 with augmentations} 

To make sure that our observed state-of-the-art distillation results are not an artifact of our well-tuned training setup, namely very long schedule and aggressive mixup augmentations, we train corresponding baseline ResNet-50 models. More specifically, we reuse the distillation training setup for supervised training on ImageNet dataset without distillation loss. To further strengthen our baseline, we additionally try SGD optimizer with momentum, which is known to often work better for ImageNet than Adam optimizer.

\begin{figure*}[t]
    \centering
    \vspace{2em}
    \begin{minipage}{0.67\textwidth}
        \centering
        \includegraphics[width=\textwidth]{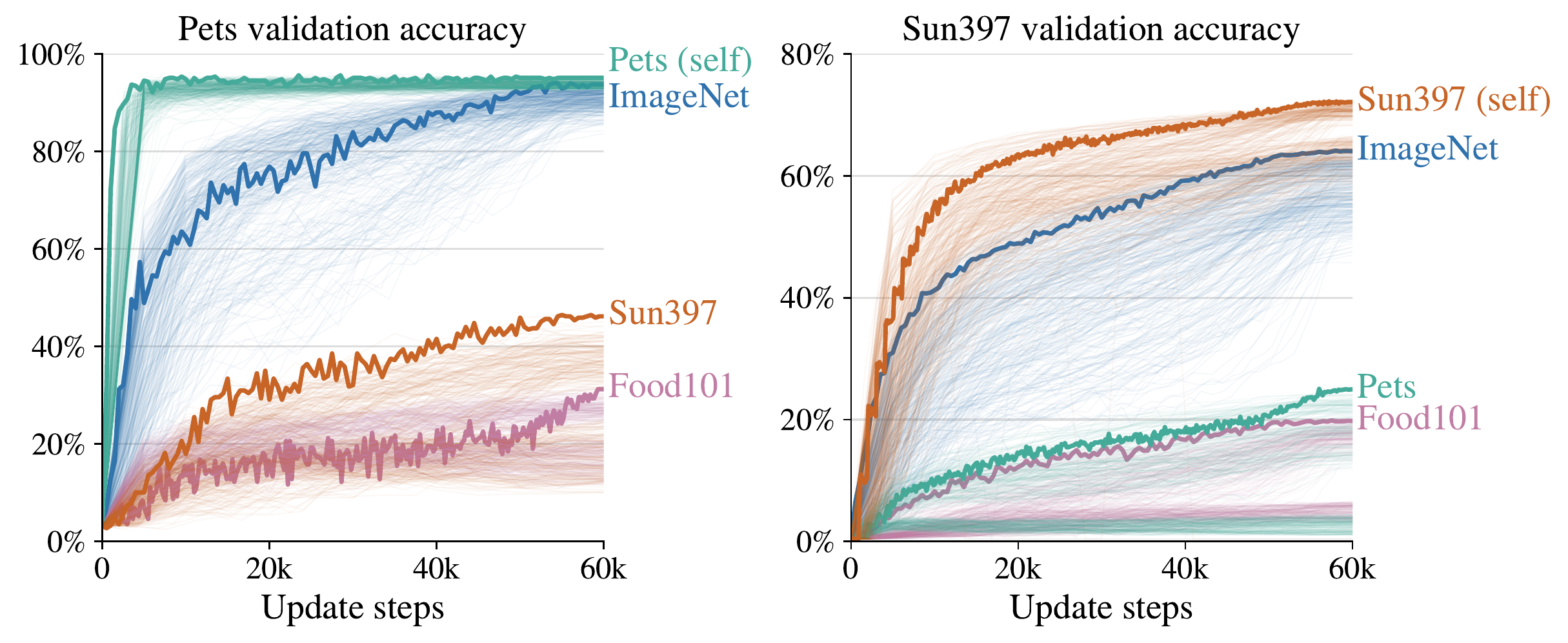}
        \captionof{figure}{Distilling \emph{pet} and \emph{sun397} datasets on different data sources. Results indicate that distilling on completely unrelated images works to some extent, even though final results are relatively low. Distilling on ``in-domain'' data is the best and distilling on related/overlapping images can work reasonably well, but may require extra long training schedule.}\label{fig:wrong_data}
    \end{minipage}
    \quad
    \begin{minipage}{0.268\textwidth}
        \centering
        \vspace{-3.81em}
        \includegraphics[width=\textwidth]{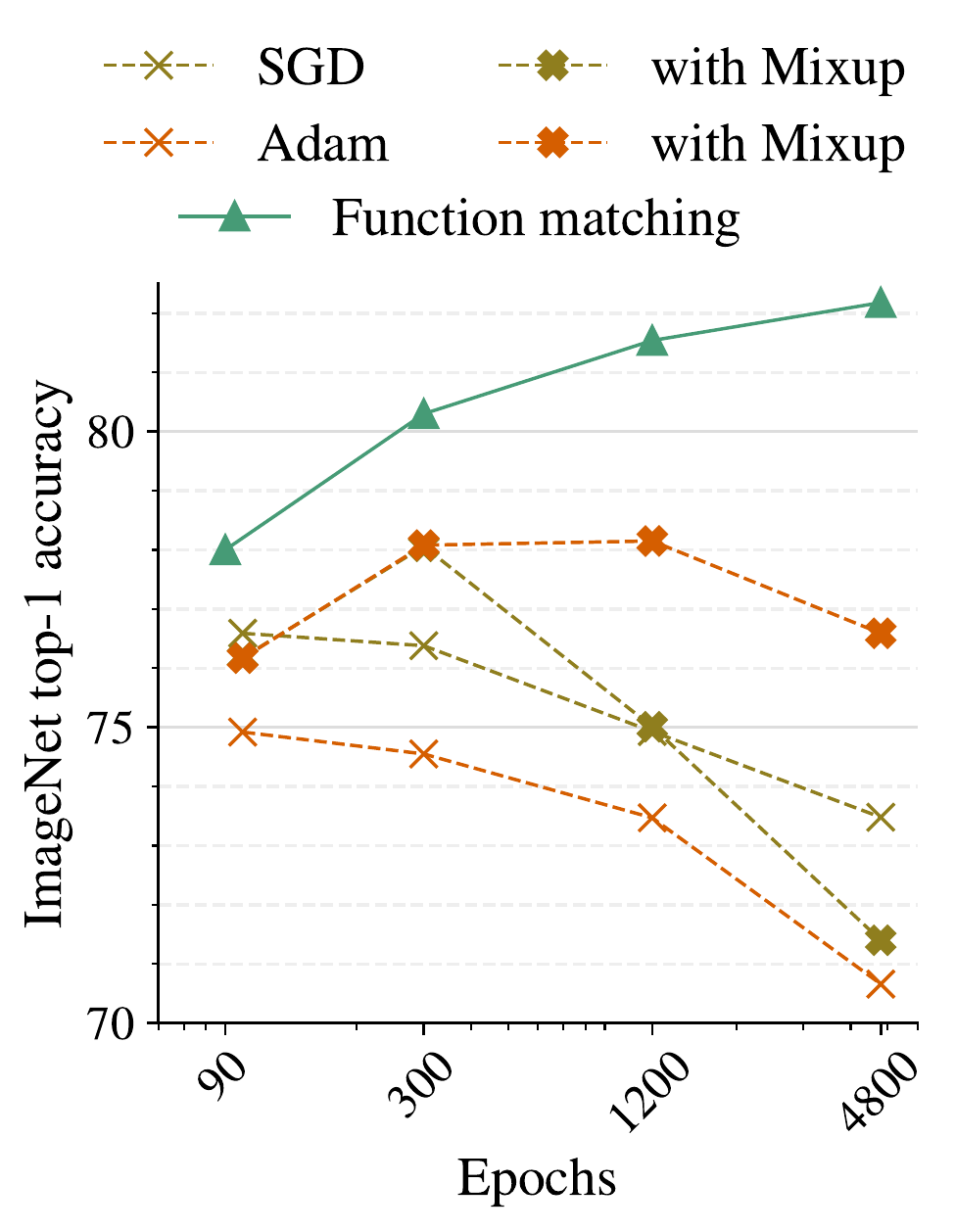}\vspace{-3mm}%
        \captionof{figure}{Baseline ResNet-50 models trained from scratch with labels vs with the ResNet-152x2 teacher.}\label{fig:imagenet_baselines}%
    \end{minipage}
\end{figure*}

Results are shown in Figure~\ref{fig:imagenet_baselines}. We observe that training with labels and without distillation loss leads to significantly worse results and starts to overfit for long training schedules. Thus, we conclude that distillation is necessary to make our training recipe work well.

\section{Related work}\label{sec:relatedwork}

There are many paradigms for compressing neural networks. One of them is \textbf{pruning}, where the general idea is to discard parts of the trained model while making it much more efficient and incurring little or no sacrifise in performance. Model pruning comes in many different flavours: it can be unstructured (\ie focus on pruning individual connections) or structured (\ie focus on pruning larger building blocks, e.g. whole channels). It can also come with or without an additional finetuning step, or be iterative or not. Balanced and fair discussion of this topic goes beyond the scope of this paper, so we refer interested reader to recent overviews as a starting point~\cite{blalock2020state,wang2021emerging}.

\textbf{Knowledge distillation}~\cite{hinton} is a technique for transferring knowledge from one model (\emph{teacher}) to another (\emph{student}), by optimizing a student model to match certain outputs (or intermediate activations) of a \emph{teacher} model. This technique is used in numerous distinct contexts, such as semi-supervised learning~\cite{tarvainen2017mean,xie2020self} or even self-supervised learning~\cite{grill2020bootstrap}. In this paper we only consider knowledge distillation as a tool for model compression.
The efficiency of distillation has been showcased in numerous works, \eg~\cite{Cho_2019_ICCV,Romero15fitnets:hints}, under different depth/width patterns of the student and teacher architectures, and even combined with other compression techniques~\cite{Mishra2018ApprenticeUK}. Notably, MEAL~\cite{shen2020mealv2} proposes to distill an ensemble of large ResNet teachers to a smaller ResNet student with an adversarial loss and achieves strong results. The main difference of our work to similar works on knowledge distillation for compression, is that our method is simultaneously the simplest and best performing: we do not introduce any new components, but rather discover that correct training setup is sufficient to attain state-of-the art results.

\textbf{Weights quantization}~\cite{lin2016fixed, jacob2018quantization, park2017weighted, wu2020integer} and \textbf{decomposition}~\cite{denton2014exploiting, lebedev2014speeding, wang2019eigendamage, gusak2019automated} aim to accelerate and reduce the memory footprint of CNNs by replacing large matrices operations with their lightweight approximations. This line of research is largely orthogonal to this work and can generally be combined with the method from this paper, especially during for the final model deployment stage. We leave exploration of this topic for future research. 

Finally, there is a line of work, which approaches our goal (compact and high performing models) from a different angle, by focusing on \textbf{altering the architecture} and getting good compact models trained from scratch, so there is no need to compress large models. Some notable examples include ResNeXt~\cite{resnext}, Squeeze-and-Excitation Networks~\cite{SEnet} and Selective Kernel~\cite{SKnet}, which propose modifications that improve model accuracy for a fixed compute budget. These improvements are complementary to the research question tackled in this paper and can be compounded.

\section{Conclusion}\label{sec:conclusion}

Instead of proposing a new method for model compression, we closely look at the existing common knowledge distillation process and identify how to make it work really well in the context of model compression. Our key findings stem from a specific interpretation of knowledge distillation: we propose to see it as a function matching task. This is not the typical view of knowledge distillation, as normally it is seen as ``a strong \emph{teacher} generates better (soft) labels that are useful for training a better and smaller \emph{student} model''.

Based on our interpretation we simultaneously incorporate three ingredients:  (i) make sure  that \emph{teacher} and \emph{student} always get identical inputs, including noise, (ii) introduce aggressive data augmentations to enrich the input image manifold (through mixup) and (iii) use very long training schedules. Even though each component of our recipe may seem trivial, our experiments show that one has to apply all of them jointly to get top results.

We attain very strong empirical results for compressing very large models to the more practical ResNet-50 architecture. We believe that they are very useful from a practical point of view and are a very strong baseline for future research on compressing large-scale models.

\textbf{Acknowledgements.} We thank Daniel Keysers and Frances Hubis for their valuable feedback on
this paper; Ilya Tolstikhin and the Google Brain team at large for providing a supportive research environment.

{\small
\bibliographystyle{ieee_fullname}
\bibliography{egbib}
}

\clearpage{}
\appendix

\section{Full results tables}\label{sec:app:table}

We provide a full summary of our experiments on ImageNet in Table~\ref{tbl:models}, with a dash ``-'' marking settings we did not deem necessary to run, as the cost outweights the potential insights.

Furthermore, Table~\ref{tbl:patience} gives numerical results for the models shown in Figure~\ref{fig:patience}, our best models (according to validation) on the four smaller datasets at $128$px resolution, together with baselines and the teacher.

\begin{table*}[t]
  \setlength{\tabcolsep}{0pt}
  \setlength{\extrarowheight}{5pt}
  \renewcommand{\arraystretch}{0.75}
  \newcolumntype{C}{>{\centering\arraybackslash}X}
  \newcolumntype{R}{>{\raggedleft\arraybackslash}X}
  \centering
  \vspace{-2em}
  \caption{Summary of all ImageNet distillation runs. Numbers represent top-1 accuracy on the validation set. By default, the student is always a ResNet50 and the teacher is BiT-M-R152x2.}\label{tbl:models}
  \begin{tabularx}{\linewidth}{p{5.4cm}p{0.1cm}Cp{0.01cm}Cp{0.01cm}Cp{0.01cm}Cp{0.01cm}Cp{0.01cm}Cp{0.01cm}C}
    \toprule[1pt]
    \bf{Experiment} && \bf{30ep} && \bf{90ep} && \bf{300ep} && \bf{600ep} && \bf{1200ep} && \bf{4800ep} && \bf{9600ep} \\
    \midrule
Best from labels     &&   -   && 76.59 && 78.08 &&   -   && 78.15 && 76.59 &&   -   \\
Fixed teacher        && 73.75 && 76.45 && 77.76 && 77.99 && 78.11 && 77.56 && 76.95 \\
consistent teacher   && 74.95 && 78.05 && 80.08 && 80.63 && 81.15 && 81.58 && 81.76 \\
function matching (FunMatch) && 73.89 && 78.00 && 80.30 && 81.17 && 81.54 && 82.18 && 82.31 \\
    \midrule
consistent teacher \shampoo     && 75.45 && 78.79 && 80.54 && 81.11 && 81.44 &&   -   &&   -   \\
function matching \shampoo       && 75.12 && 78.70 && 80.63 &&   -   && 81.67 &&   -   &&   -   \\
    \midrule
T224 $\rightarrow$ S160 (consistent teacher)  && 71.38 && 75.57 && 78.01 &&   -   &&   -   &&   -   &&   -   \\
T224 $\rightarrow$ S160 (function matching)   && 70.22 && 75.34 && 78.17 && 79.07 && 79.61 &&  0.10 && 80.49 \\
    \arrayrulecolor{lightgray}\midrule[0.25pt]\arrayrulecolor{black}
FunMatch: T384 $\rightarrow$ S224          &&   -   &&   -   && 80.46 &&   -   && 81.82 && 82.33 && 82.64 \\
    \arrayrulecolor{lightgray}\midrule[0.25pt]\arrayrulecolor{black}
FunMatch: T384+224 $\rightarrow$ S224      &&   -   &&   -   &&   -   &&   -   && 82.12 && 82.71 && 82.82 \\
    \midrule
FunMatch: MobileNet v3 (GN)    &&   -   &&   -   && 74.60 &&   -   && 76.31 && 76.84 && 76.97 \\
FunMatch: MobileNet v3 (GN, 2T) &&   -   &&   -   && 74.85 &&   -   && 76.51 &&   -   &&   -   \\
FunMatch: MobileNet v3 (GN, Small) &&   -   &&   -   && 65.61 &&   -   && 67.57 &&   -   &&   -   \\
    \arrayrulecolor{lightgray}\midrule[0.25pt]\arrayrulecolor{black}
FunMatch: MobileNet v3 (BN)    &&   -   &&   -   && 72.32 &&   -   && 73.51 &&   -   &&   -   \\
FunMatch: MobileNet v3 (BN, 2T) &&   -   &&   -   && 73.28 &&   -   &&   -   &&   -   &&   -   \\
    \midrule
Figure~\ref{fig:imagenet} (right): BiT-M-R50 init       && 77.52 && 79.43 && 80.47 && 80.83 && 81.11 && 81.45 &&   -   \\
Figure~\ref{fig:imagenet_baselines}: SGDM && - && 76.59 && 76.38 && - && 74.93 && 73.48 && - \\
Figure~\ref{fig:imagenet_baselines}: Adam && - && 74.92 && 74.55 && - && 73.47 && 70.66 && - \\
Figure~\ref{fig:imagenet_baselines}: SGDM + Mixup && - && 76.18 && 78.06 && - && 75.01 && 71.40 && - \\
Figure~\ref{fig:imagenet_baselines}: Adam + Mixup && - && 76.17 && 78.08 && - && 78.15 && 76.59 && - \\
    \bottomrule
  \end{tabularx}
\end{table*}

\begin{table*}[t]
  \setlength{\tabcolsep}{5pt}
  \setlength{\extrarowheight}{5pt}
  \renewcommand{\arraystretch}{0.75}
  \centering
  \caption{Tabular representation of the results from Figure~\ref{fig:patience}.}\label{tbl:patience}
  \begin{tabulary}{1.0\textwidth}{LRCRLL}
    \toprule[1pt]
    \bf{Model} & \centering \bf{Epochs} & \bf{Final Test Acc} & \centering \bf{T} & \bf{LR} & \bf{WD} \\
\midrule
\midrule
\multicolumn{6}{c}{\emph{Flowers102}} \\
\arrayrulecolor{lightgray}\midrule[0.25pt]\arrayrulecolor{black}
ResNet-50x1 student        &        1000 & 77.51\% & 10 &   0.003 & 0.001 \\
ResNet-50x1 student        &     10\,000 & 92.83\% & 10 &   0.003 & 0.0003 \\
ResNet-50x1 student        &    100\,000 & 95.54\% &  1 &   0.001 & 0.0001 \\
ResNet-50x1 student        & 1\,000\,000 & 96.93\% &  1 &  0.0003 & 1e-05 \\
\arrayrulecolor{lightgray}\midrule[0.25pt]\arrayrulecolor{black}
ResNet-152x2 teacher       & - & 97.82\% & - & - & - \\
Best transfer ResNet50     &     10\,000 & 97.50\% & \multicolumn{3}{c}{LR=0.01, Mixup=0.0}  \\
Best from-scratch ResNet50 &     10\,000 & 66.38\% & \multicolumn{3}{c}{LR=0.01, Mixup=1.0}  \\
\midrule
\multicolumn{6}{c}{\emph{Pet37}} \\
\arrayrulecolor{lightgray}\midrule[0.25pt]\arrayrulecolor{black}
ResNet-50x1 student        &         300 & 82.75\% &  2 &    0.01 & 1e-05 \\
ResNet-50x1 student        &        1000 & 88.01\% &  5 &    0.01 & 0.001 \\
ResNet-50x1 student        &        3000 & 90.08\% & 10 &   0.003 & 0.0003 \\
ResNet-50x1 student        &     10\,000 & 90.98\% &  2 &   0.001 & 0.0001 \\
ResNet-50x1 student        &     30\,000 & 91.06\% &  2 &   0.003 & 1e-05 \\
\arrayrulecolor{lightgray}\midrule[0.25pt]\arrayrulecolor{black}
ResNet-152x2 teacher       & - & 91.03\% & - & - & - \\
Best transfer ResNet50     &     10\,000 & 88.20\% & \multicolumn{3}{c}{LR=0.001, Mixup=1.0}  \\
Best from-scratch ResNet50 &     10\,000 & 74.24\% & \multicolumn{3}{c}{LR=0.01, Mixup=1.0}  \\
\midrule
\multicolumn{6}{c}{\emph{Food101}} \\
\arrayrulecolor{lightgray}\midrule[0.25pt]\arrayrulecolor{black}
ResNet-50x1 student        &         100 & 83.29\% & 10 &    0.01 & 0.001 \\
ResNet-50x1 student        &        1000 & 86.64\% & 10 &   0.001 & 0.0003 \\
ResNet-50x1 student        &        3000 & 87.20\% &  5 &    0.01 & 0.0001 \\
\arrayrulecolor{lightgray}\midrule[0.25pt]\arrayrulecolor{black}
ResNet-152x2 teacher       & - & 86.24\% & - & - & - \\
Best transfer ResNet50     &        1000 & 85.05\% & \multicolumn{3}{c}{LR=0.001, Mixup=1.0}  \\
Best from-scratch ResNet50 &        1000 & 74.56\% & \multicolumn{3}{c}{LR=0.01, Mixup=1.0}  \\
\midrule
\multicolumn{6}{c}{\emph{Sun397}} \\
\arrayrulecolor{lightgray}\midrule[0.25pt]\arrayrulecolor{black}
ResNet-50x1 student        &         100 & 68.28\% & 10 &    0.01 & 0.001 \\
ResNet-50x1 student        &        1000 & 73.46\% & 10 &   0.003 & 0.0001 \\
ResNet-50x1 student        &        3000 & 74.26\% & 10 &    0.01 & 3e-05 \\
\arrayrulecolor{lightgray}\midrule[0.25pt]\arrayrulecolor{black}
ResNet-152x2 teacher       & - & 74.22\% & - & - & - \\
Best transfer ResNet50     &        1000 & 71.61\% & \multicolumn{3}{c}{LR=0.001, Mixup=1.0} \\
Best from-scratch ResNet50 &        1000 & 60.63\% & \multicolumn{3}{c}{LR=0.01, Mixup=1.0}  \\
    \bottomrule
  \end{tabulary}
\end{table*}

\section{BiT models download statistics}\label{sec:app:download}

\begin{figure*}[ht]%
    \centering
    \includegraphics[width=\textwidth]{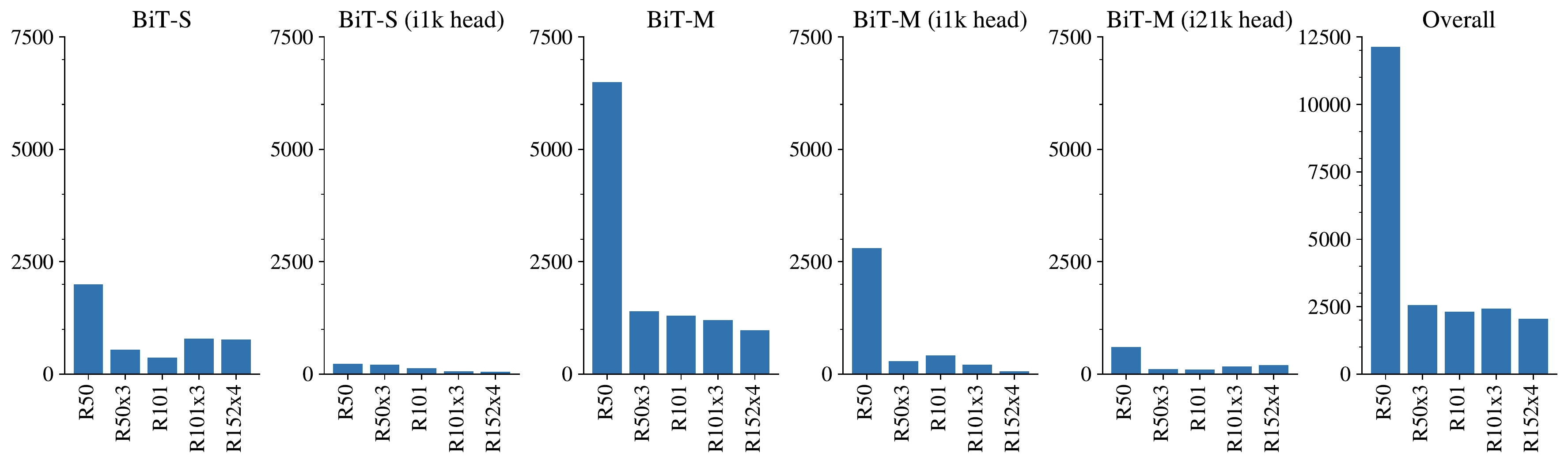}
    \caption{
      BiT models download statistics according to \url{https://tfhub.dev/google/collections/bit}.
      ``BiT-S''/``BiT-M'' denotes the BiT model for feature extraction, 
      while the figures with a mention of ``head'' correspond to the classifiers.
      The rightmost overall plot shows the total download counts for each size. 
      It is clear that ResNet-50 is by far the most widely used model.
    }%
    \label{fig:app:download}%
\end{figure*}

In Figure~\ref{fig:app:download}, we show the download statistics for models with different sizes: ResNet50, ResNet50x3, ResNet101, ResNet101x3 and ResNet152x4. 
It's clear that the smallest ResNet50 model is the most used, with a significant gap compared to the other models.
The practitioners' behavior motivates our work of getting the best possible ResNet50 model.

\section{More consistency plots}\label{sec:app:consistency}

\begin{figure*}[ht]%
    \centering
    \includegraphics[width=\textwidth]{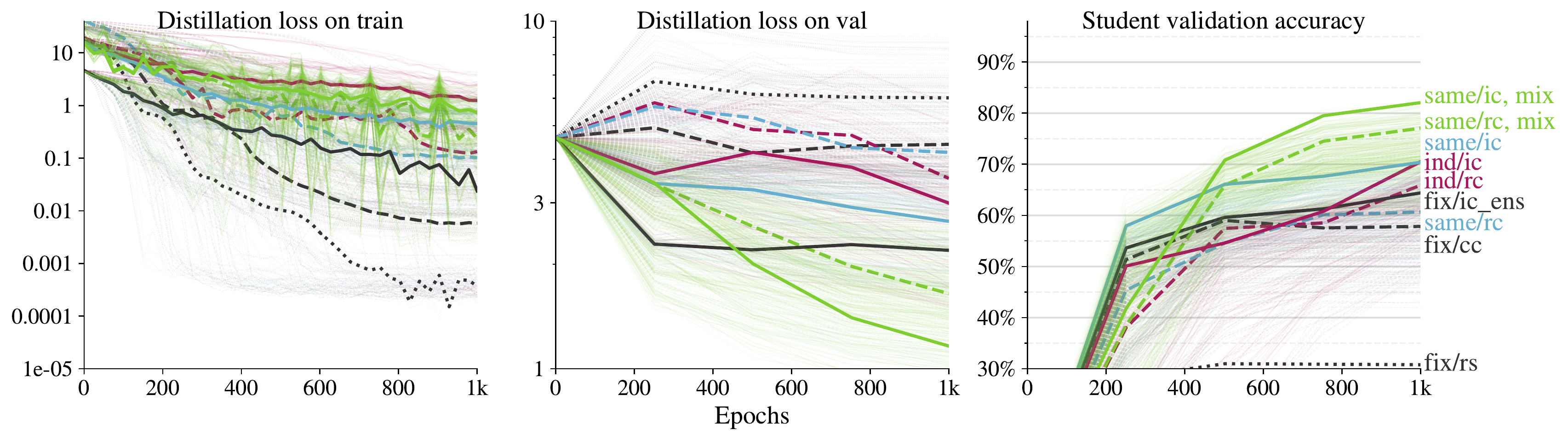}
    \includegraphics[width=\textwidth]{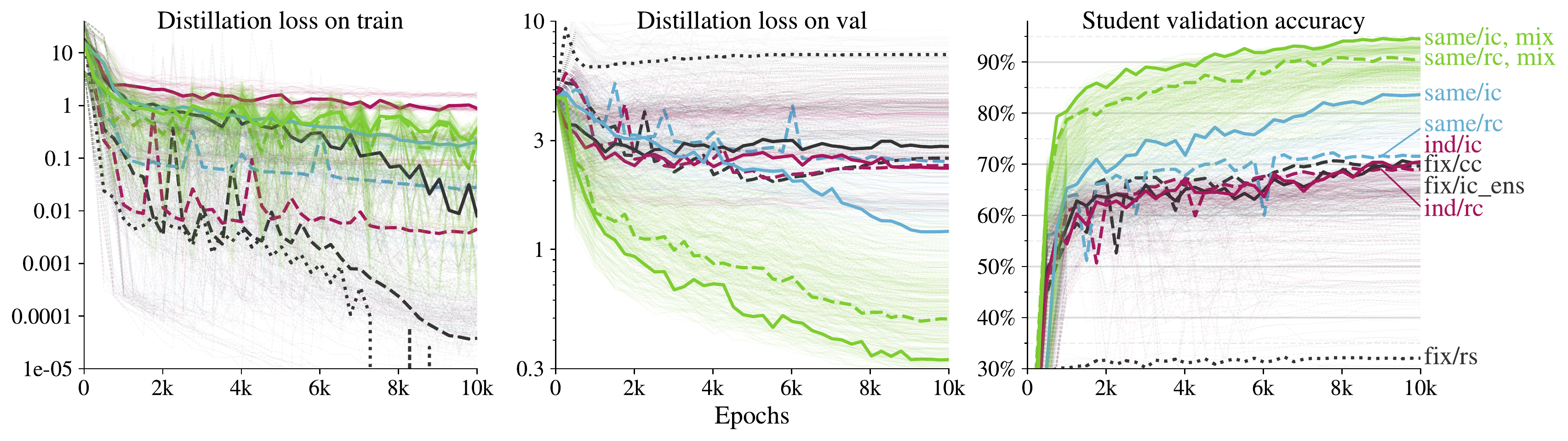}
    \includegraphics[width=\textwidth]{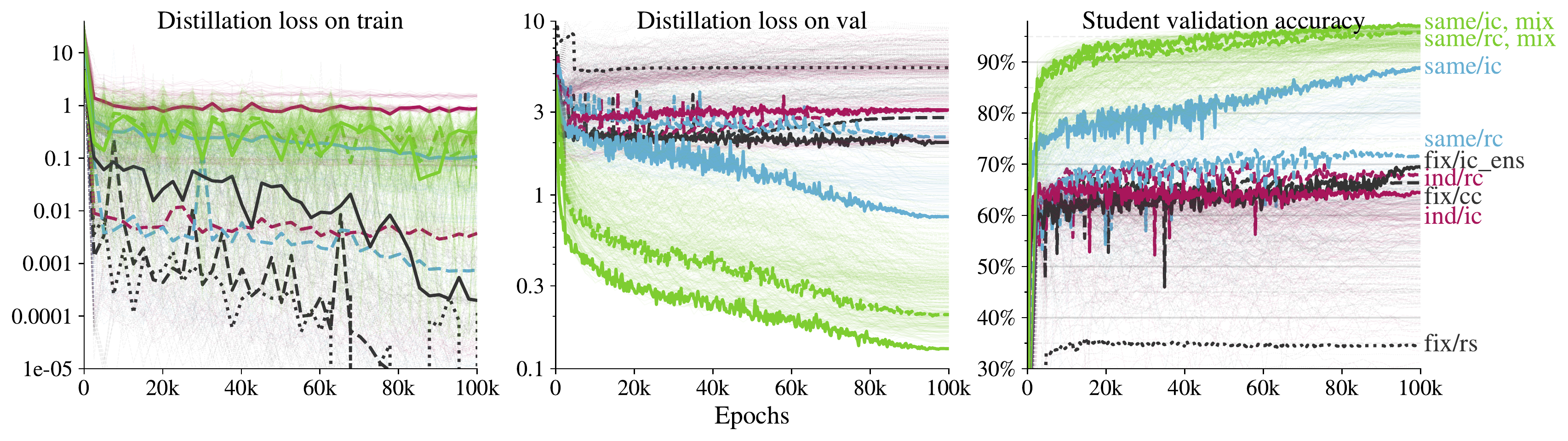}
    \caption{Consistency plots for the Flowers102 dataset, when training for 1\,000~epochs, 10\,000~epochs, and 100\,000~epochs, from top to bottom respectively.}%
    \label{fig:app:consistency_flowers}%
\end{figure*}

\begin{figure*}[ht]%
    \centering
    \includegraphics[width=\textwidth]{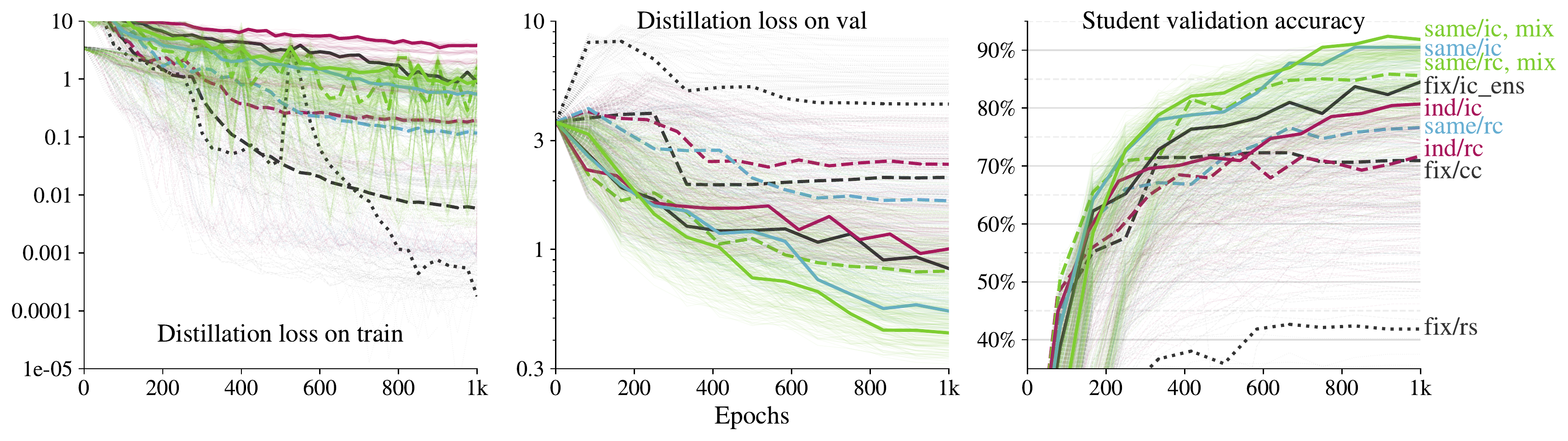}
    \includegraphics[width=\textwidth]{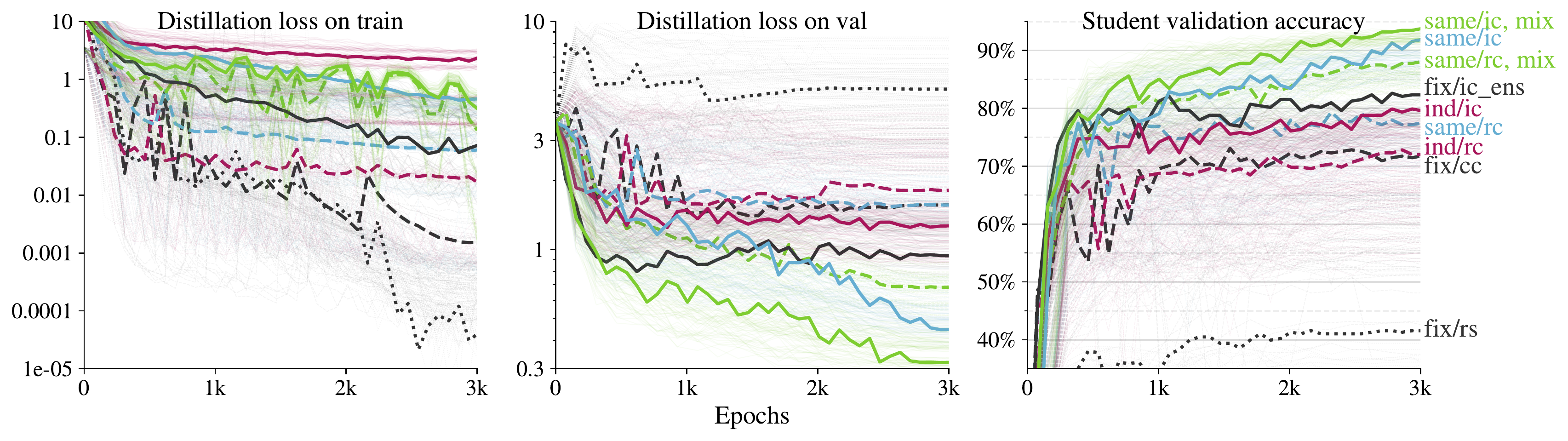}
    \includegraphics[width=\textwidth]{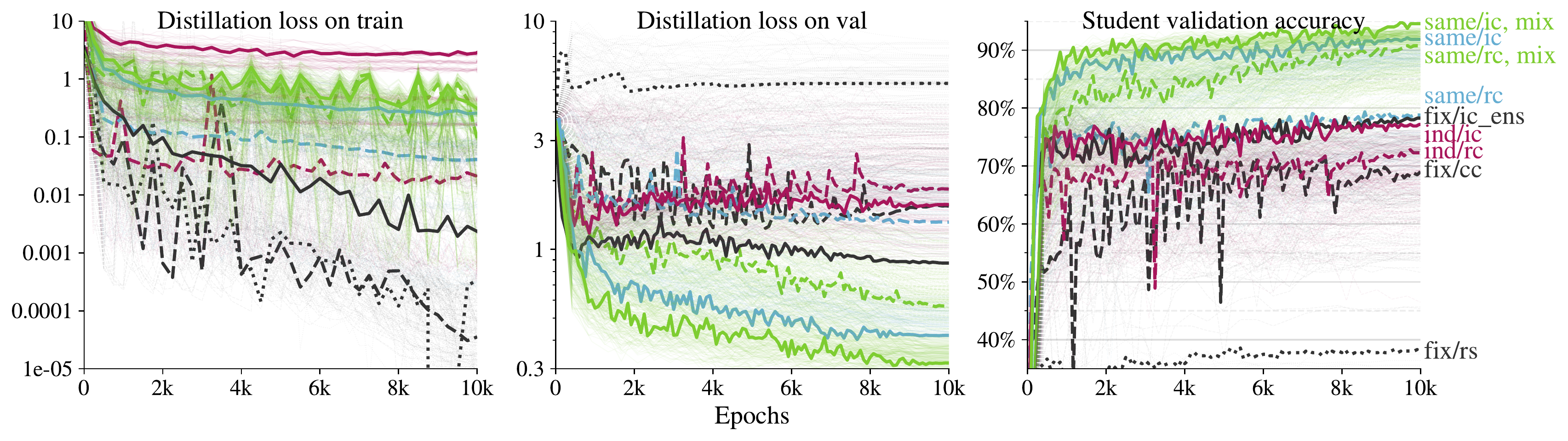}
    \includegraphics[width=\textwidth]{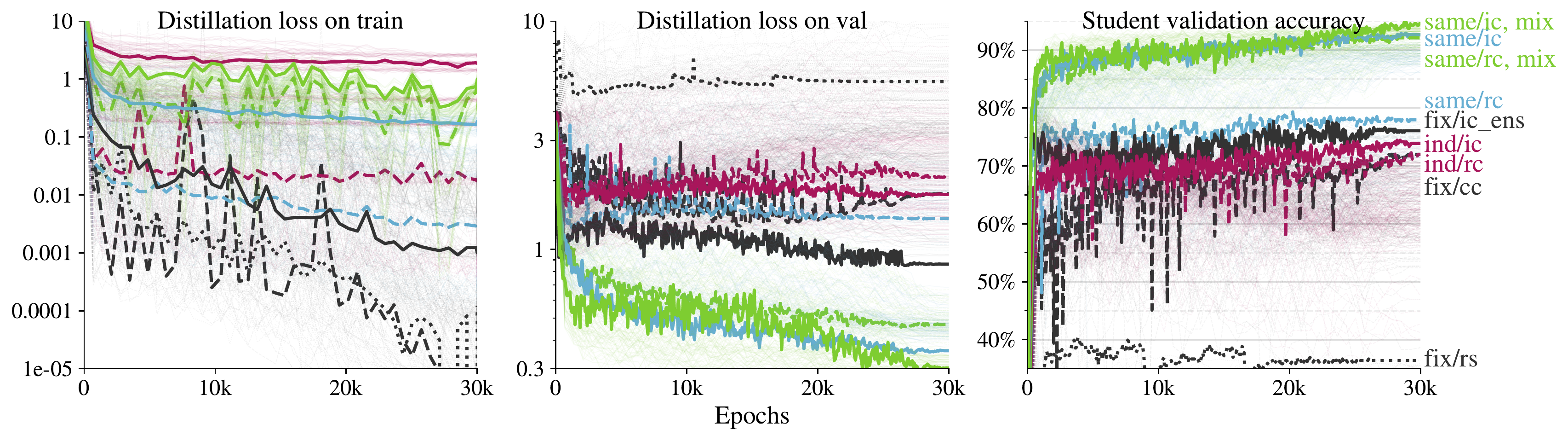}
    \caption{Consistency plots for the Pet37 dataset, when training for 1\,000~epochs, 3\,000~epochs, 10\,000~epochs, and 30\,000~epochs, from top to bottom respectively.}%
    \label{fig:app:consistency_pets}%
\end{figure*}

\begin{figure*}[ht]%
    \centering
    \includegraphics[width=\textwidth]{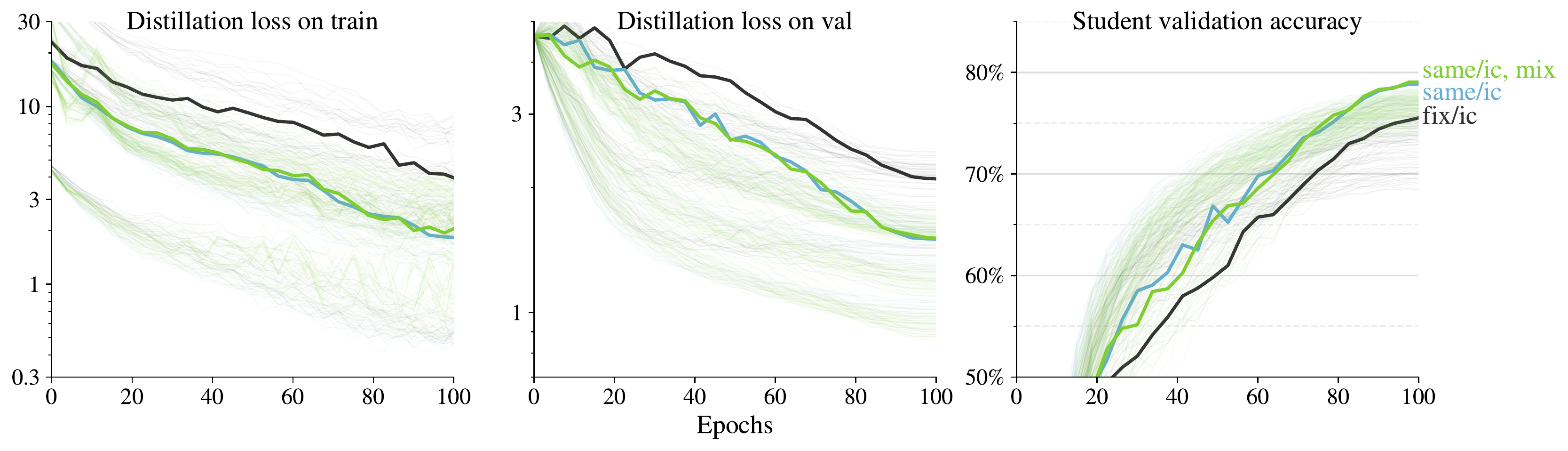}
    \includegraphics[width=\textwidth]{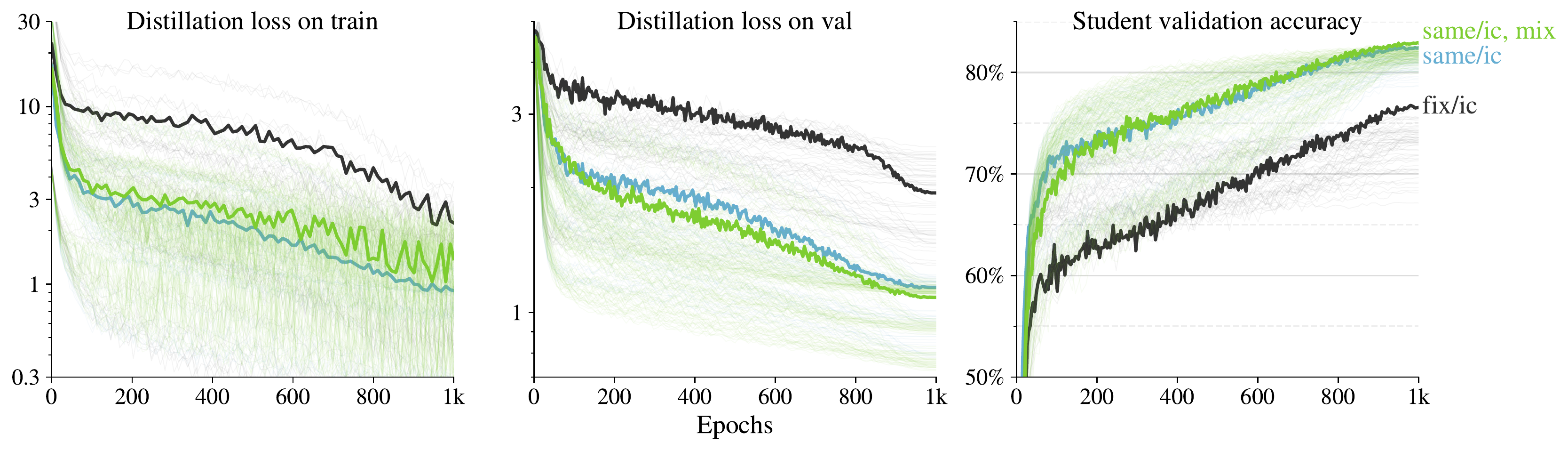}
    \includegraphics[width=\textwidth]{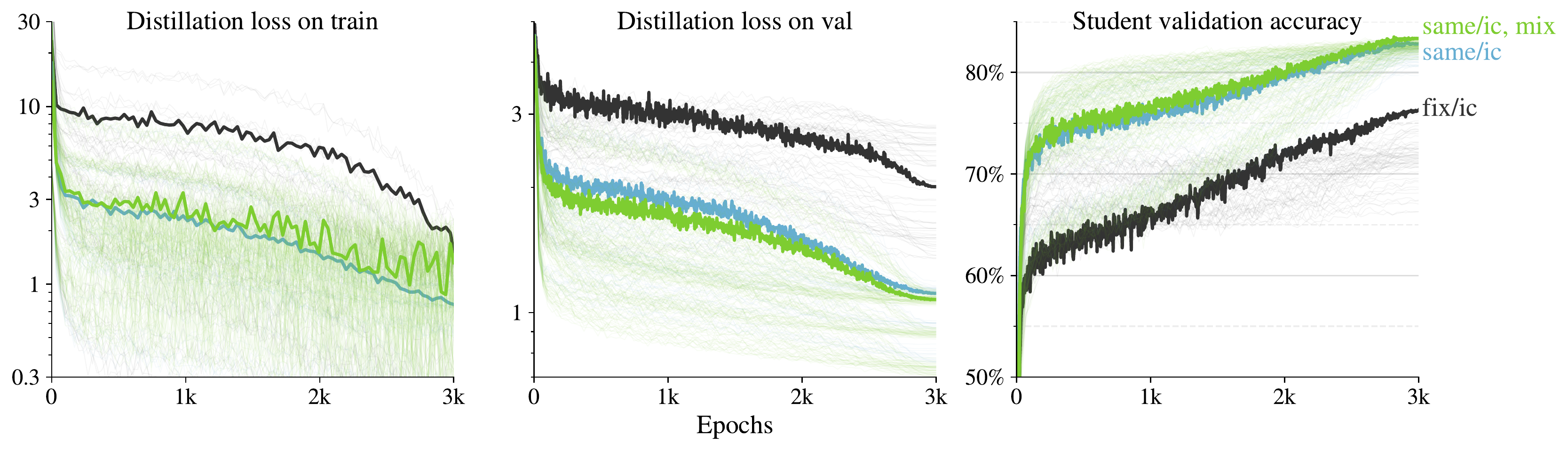}
    \caption{Consistency plots for the Food101 dataset, when training for 100~epochs, 1\,000~epochs, and 3\,000~epochs, from top to bottom respectively.}%
    \label{fig:app:consistency_food}%
\end{figure*}

\begin{figure*}[ht]%
    \centering
    \includegraphics[width=\textwidth]{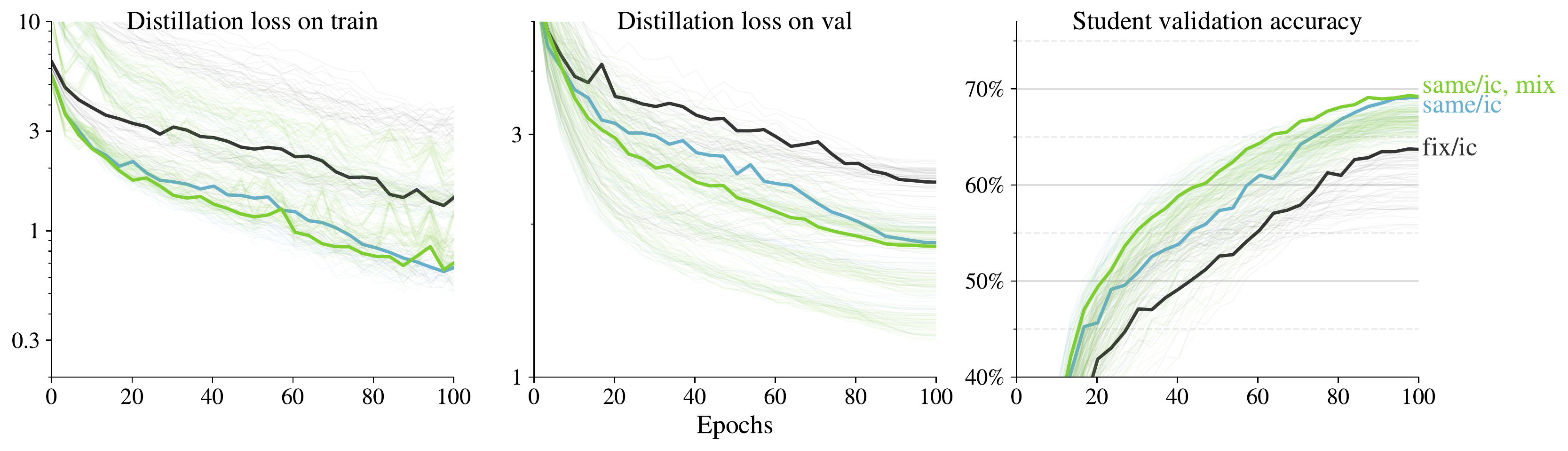}
    \includegraphics[width=\textwidth]{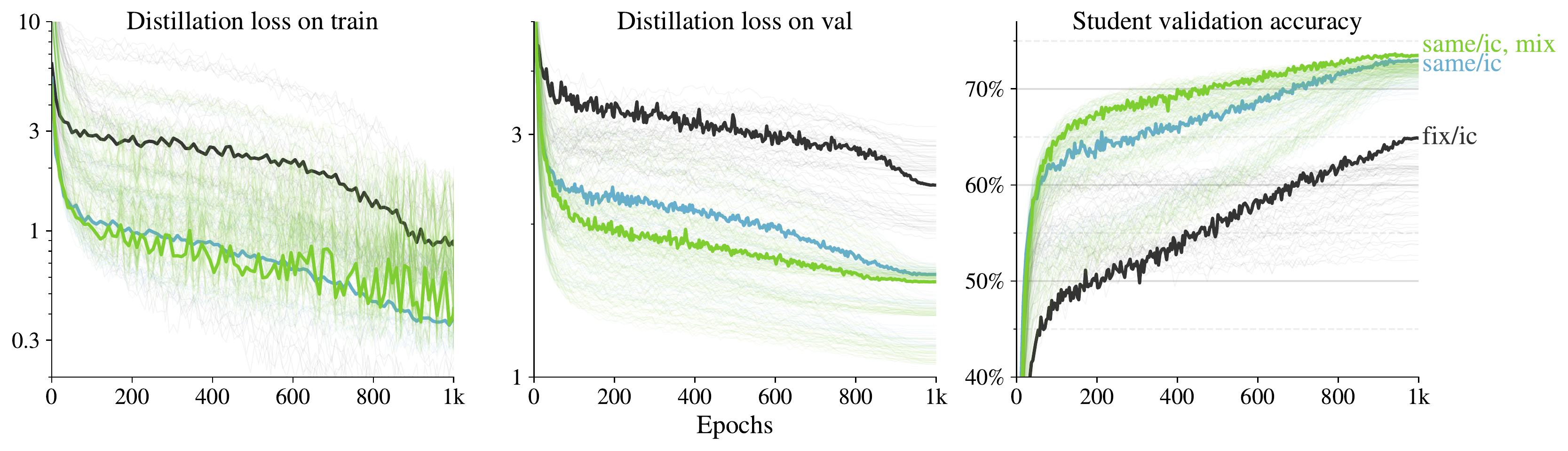}
    \includegraphics[width=\textwidth]{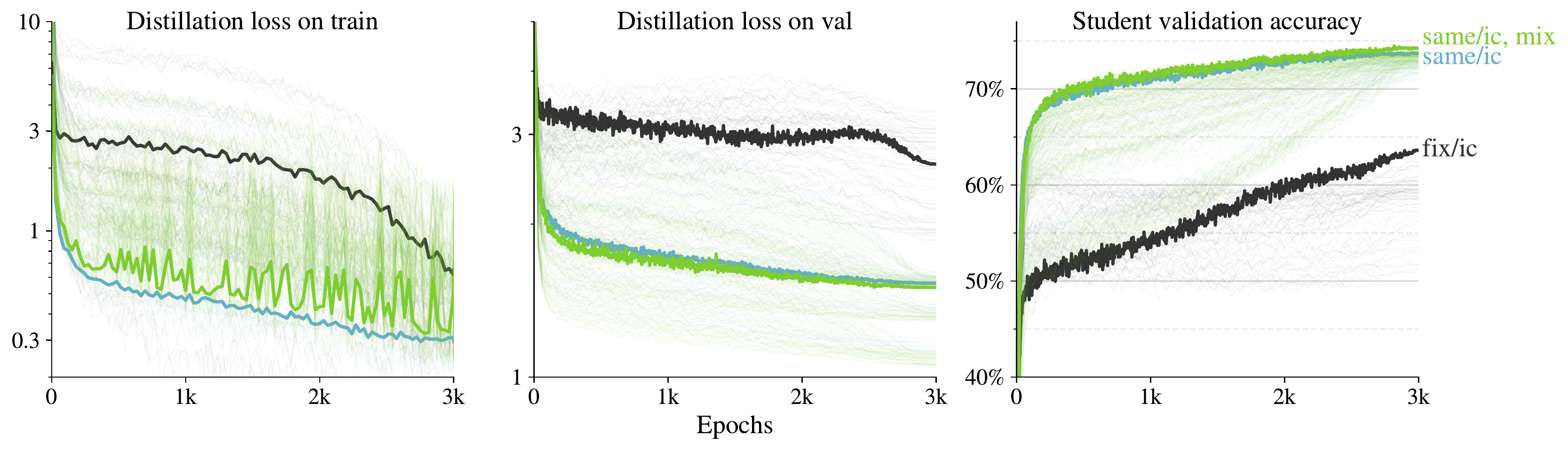}
    \caption{Consistency plots for the SUN397 dataset, when training for 100~epochs, 1\,000~epochs, and 3\,000~epochs, from top to bottom respectively.}%
    \label{fig:app:consistency_sun}%
\end{figure*}

In Figures~\ref{fig:app:consistency_flowers}~to~\ref{fig:app:consistency_sun}, we show the ``consistency'' plots (cf Figure~\ref{fig:consistency} in the main paper) for all datasets and across all training durations.
It is noteworthy that (relatively) short runs may provide deceptive signal on the best method, and only with the addition of ``patience'', \eg when distilling for a long time, does it become clear that the full function-matching approach is the best choice.
\section{Shampoo optimization details}\label{sec:app:shampoo}
For all experiments the learning rate schedule was a linear warm-up up to 1800 steps followed by a quadratic decay towards zero. Overhead of Shampoo is quite minimal due blocking trick (each preconditioner is atmost 128x128) and inverse is run in a distributed manner across the TPU cores every step, with nesterov momentum. These settings are identical to the the training recipe in \cite{anil2021scalable} for training a ResNet-50 architecture on ImageNet from scratch efficiently at large batch sizes.  All experiments uses weight decay of 0.000375. 

\section{Training, validation and test splits}\label{app:splits}

\begin{table*}[t]
  \setlength{\tabcolsep}{0pt}
  \setlength{\extrarowheight}{5pt}
  \renewcommand{\arraystretch}{0.75}
  \newcolumntype{C}{>{\centering\arraybackslash}X}
  \newcolumntype{R}{>{\raggedleft\arraybackslash}X}
  \centering
  \caption{\emph{Train}, \emph{validation} and \emph{test} splits. Split definitions follow notation from the \emph{tensorflow datasets} library and can be directly used to access relevant data splits using the library.}\label{tbl:splits}
  \vspace{1mm}
  \begin{tabularx}{\linewidth}{p{2.3cm}p{0.1cm}Cp{0.01cm}Cp{0.01cm}Cp{0.01cm}C}
    \toprule[1pt]
    \bf{Dataset} && \bf{train split} && \bf{validation split} && \bf{test split} \\
    \midrule
    Flowers102 && \texttt{train} && \texttt{validation} && \texttt{test} \\
    Pets37 &&  \texttt{train[:90}\%\texttt{]} &&  \texttt{train[90}\%\texttt{:]} &&  \texttt{test} \\
    Food101 &&  \texttt{train[:90}\%\texttt{]} &&  \texttt{train[90}\%\texttt{:]} &&  \texttt{test} \\
    Sun397 && \texttt{train} && \texttt{validation} && \texttt{test} \\
    ImageNet &&  \texttt{train[:98}\%\texttt{]} &&  \texttt{train[98}\%\texttt{:]} &&  \texttt{validation} \\
    \bottomrule
  \end{tabularx}
\end{table*}

Throughout our experiments we rely on the \emph{tensorflow datasets} library\footnote{\url{https://www.tensorflow.org/datasets}} to access all datasets. A huge advantage of this library is that it enables a unified and reproducible way to access diverse datasets. To this end, we report our \emph{train}, \emph{validation} and \emph{test} splits (following the library's notation) in Table~\ref{tbl:splits}.

\section{Configuration file for ImageNet distillation}\label{app:configs}

We present the configuration for performing distillation on ImageNet following the \texttt{big\_vision}~\cite{big_vision} conventions.

\begin{lstlisting}[language=Python, caption=Full config for ImageNet distillation.]
def get_config():
  config = mlc.ConfigDict()

  config.dataset = 'imagenet2012'
  config.train_split = 'train[:98%]'
  config.num_classes = 1000

  config.batch_size = 4096
  config.num_epochs = 1200
  config.shuffle_buffer_size = 250_000

  config.log_training_steps = 50
  config.checkpoint_steps = 1000
  config.keep_checkpoint_steps = 20000

  # Model section
  config.student_name = 'bit_paper'
  config.student = dict(depth=50, width=1)

  config.teachers = ['prof_m']
  config.prof_m_name = 'bit_paper'
  config.prof_m_init = 'FILENAME'
  config.prof_m = dict(depth=152, width=2)

  pp_common = (
    '|value_range(-1, 1)'
    '|onehot(1000, key="{lbl}", key_result="labels")'
    '|keep("image", "labels")'
  )
  config.pp_train = (
    'decode_jpeg_and_inception_crop(224)|flip_lr' 
    + pp_common.format(lbl='label')
  )
  ppv = 'decode|{crop}' + pp_common

  config.mixup = dict(p=1.0, n=2)

  # Distillation settings
  config.distance = 'kl'
  config.distance_kw = dict(t=1.0)

  # Optimizer section
  config.grad_clip_norm = 1.0
  config.optax_name = 'scale_by_adam'
  config.optax = dict(mu_dtype='bfloat16')

  config.lr = 0.03
  config.wd = 0.0003
  config.schedule = dict(warmup_steps=5000, decay_type='cosine')

  # Eval section shortened for brevity,
  # see code release for full details.
  minitrain_split = 'train[:2%]'
  minival_split = 'train[99%:]'
  val_split = 'validation'

  return config

\end{lstlisting}

\end{document}